\documentclass[]{IEEEtran}
\usepackage[table]{xcolor}
\usepackage{cite}

\usepackage{graphicx}
\usepackage{amsmath}
\usepackage{flushend}
\usepackage{subfigure}
\usepackage{amssymb}
\usepackage{booktabs}
\usepackage{fancybox}
\usepackage{tikz}
\usepackage{array}
\usepackage{bbding}
\usepackage{amssymb}
\usepackage{subfigure}
\usepackage{makecell}
\makeatletter
\let\NAT@parse\undefined
\makeatother

\definecolor{tsneRed}{rgb}{0.85, 0.13, 0.16}

\definecolor{tsneGreen}{rgb}{0.00, 0.59, 0.20}

\usepackage{tikz}
\usepackage{hyperref}  
\ifCLASSINFOpdf
\else
\fi
\usepackage{multirow} 
\definecolor{lime}{HTML}{A6CE39}
\DeclareRobustCommand{\orcidicon}{%
    \begin{tikzpicture}
    \draw[lime, fill=lime] (0,0) 
    circle [radius=0.16] 
    node[white] {{\fontfamily{qag}\selectfont \tiny ID}};    \draw[white, fill=white] (-0.0625,0.095) 
    circle [radius=0.007];    \end{tikzpicture}
    \hspace{-2mm}}
\foreach \x in {A, ..., Z}{%
    \expandafter\xdef\csname orcid\x\endcsname{\noexpand\href{https://orcid.org/\csname orcidauthor\x\endcsname}{\noexpand\orcidicon}}
    }

\begin{document}

%
\title{Multi-Stage Knowledge Integration of Vision-Language Models for Continual Learning}
%
%
%
\newcommand{\orcidauthorA}{0000-0002-2197-3739}
\newcommand{\orcidauthorB}{0000-0001-6670-3727}
\newcommand{\orcidauthorC}{0000-0003-4361-956X}
\newcommand{\orcidauthorD}{0009-0009-0163-4105}
\author{Hongsheng~Zhang,
        Zhong~Ji\orcidA{}, \textit{Senior Member, IEEE},
        Jingren~Liu,
        Yanwei~Pang\orcidB{}, \textit{Senior Member, IEEE},
        and~Jungong~Han\orcidC{}, \textit{Senior Member, IEEE}

\thanks{Manuscript received xxx.}
\thanks{This work was supported in part by the National Key Research and Development Program of China under Grant 2022ZD0160403; in part by the National Natural Science Foundation of China (NSFC) under Grants 62176178 and 62106152.}
\thanks{Hongsheng Zhang, Zhong Ji, Jingren Liu, and Yanwei Pang are with the School of Electrical and Information Engineering and the Tianjin Key Laboratory of Brain-Inspired Intelligence Technology, Tianjin University, Tianjin 300072, China and also with the Shanghai Artificial Intelligence Laboratory, Shanghai 200232 (e-mail: zhs0204@tju.edu.cn; jizhong@tju.edu.cn; jrl0219@tju.edu.cn; pyw@tju.edu.cn).}
\thanks{Jungong Han is with the Department of Computer Science, The University of Sheffield, Sheffield, U.K. (e-mail: jungonghan77@gmail.com).}
\thanks{The corresponding author is Zhong Ji.}

}

\markboth{Submitted to IEEE TRANSACTIONS ON IMAGE PROCESSING,~Vol.~XX, No.~X, July~20XX}%
{Zhang \MakeLowercase{\textit{et al.}}: Multi-Stage Knowledge Integration of Vision-Language Models for Continual Learning}

\maketitle

\begin{abstract}
Vision Language Models (VLMs), pre-trained on large-scale image-text datasets, enable zero-shot predictions for unseen data but may underperform on specific unseen tasks. Continual learning (CL) can help VLMs effectively adapt to new data distributions without joint training, but faces challenges of catastrophic forgetting and generalization forgetting. 
Although significant progress has been achieved by distillation-based methods, they exhibit two severe limitations. One is the popularly adopted single-teacher paradigm fails to impart comprehensive knowledge, The other is the existing methods inadequately leverage the multimodal information in the original training dataset, instead they rely on additional data for distillation, which increases computational and storage overhead. 
To mitigate both limitations, by drawing on Knowledge Integration Theory (KIT), we propose a Multi-Stage Knowledge Integration network (MulKI) to emulate the human learning process in distillation methods. MulKI achieves this through four stages, including Eliciting Ideas, Adding New Ideas, Distinguishing Ideas, and Making Connections. 
During the four stages, we first leverage prototypes to align across modalities, eliciting cross-modal knowledge, then adding new knowledge by constructing fine-grained intra- and inter-modality relationships with prototypes. After that, knowledge from two teacher models is adaptively distinguished and re-weighted. Finally, we connect between models from intra- and inter-task, integrating preceding and new knowledge.
Our method demonstrates significant improvements in maintaining zero-shot capabilities while supporting continual learning across diverse downstream tasks, showcasing its potential in adapting VLMs to evolving data distributions.
\end{abstract}


\IEEEpeerreviewmaketitle

\section{Introduction}

 \begin{figure}[htbp]
    \begin{center}
	\includegraphics[scale=0.46]{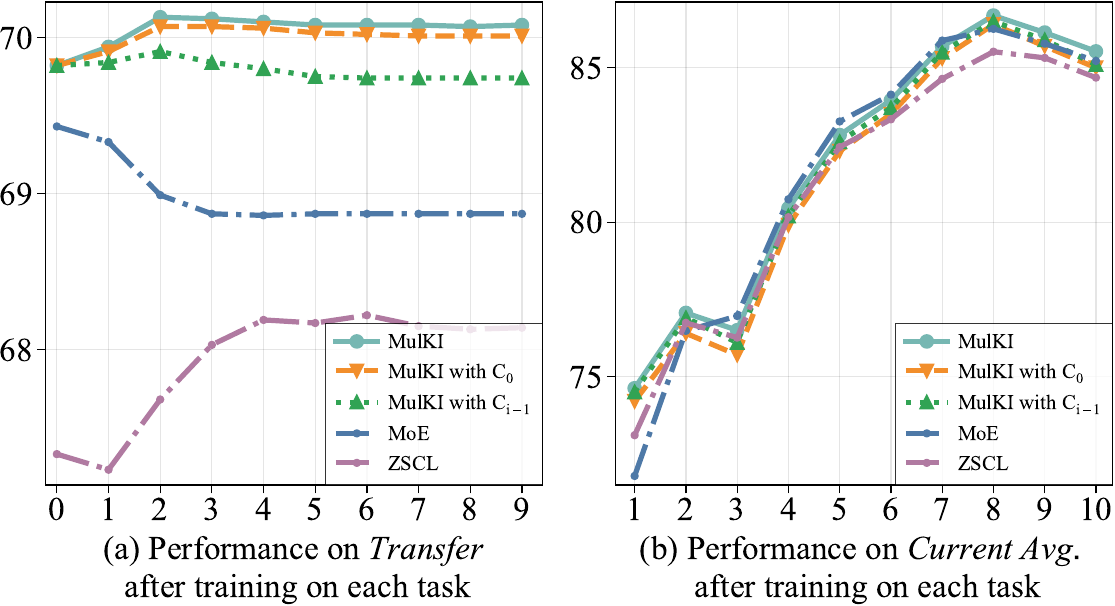}
	\end{center}

      \caption{ Performance (\%) changes during continual learning on MTIL\cite{zscl} for each task, in which MulKI with $C_0$ and with $C_{i-1}$ represent adopting only the initial model and only the preceding model for distillation, respectively. The metric ``Transfer" denotes averaging the results of unseen tasks only, while ``Current Avg." averages only the results of seen tasks, the two metrics represent the mitigation of two forgetting issues, respectively. Our method effectively mitigates both forgetting issues. We omit the results of task 0 (lower than 55\%) in (b) for clearer observation. We exclude LwF-VR here for its low accuracy.}\label{tu1}
\end{figure}
\IEEEPARstart{I}{n} recent years, Vision Language Models (VLMs) have garnered substantial attention due to their exceptional zero-shot generalization capabilities \cite{jia2021scaling, li2022blip, yao2021filip, zy}, which is largely attributed to their extensive training on vast datasets. However, in real-world applications, data typically arrives in a continuous, streaming fashion, rendering large-scale, unified training less feasible. 
As a result, Continual Learning techniques for Vision Language Models (CL-VLMs) have been developed to focus on the incremental and adaptive learning for VLMs. Nevertheless, during the adaptation process for downstream tasks, CL-VLMs encounter two predominant challenges: 1) catastrophic forgetting\cite{catas}, which entails the erosion of previously acquired knowledge, and 2) generalization forgetting, which compromises the model's zero-shot generalization capability. These challenges pose significant obstacles to the practical deployment of VLMs, underscoring the need for advanced strategies to mitigate the adverse effects of knowledge degradation across evolving data streams.\par

\begin{figure*}
	\begin{center}
		\includegraphics[scale=0.5]{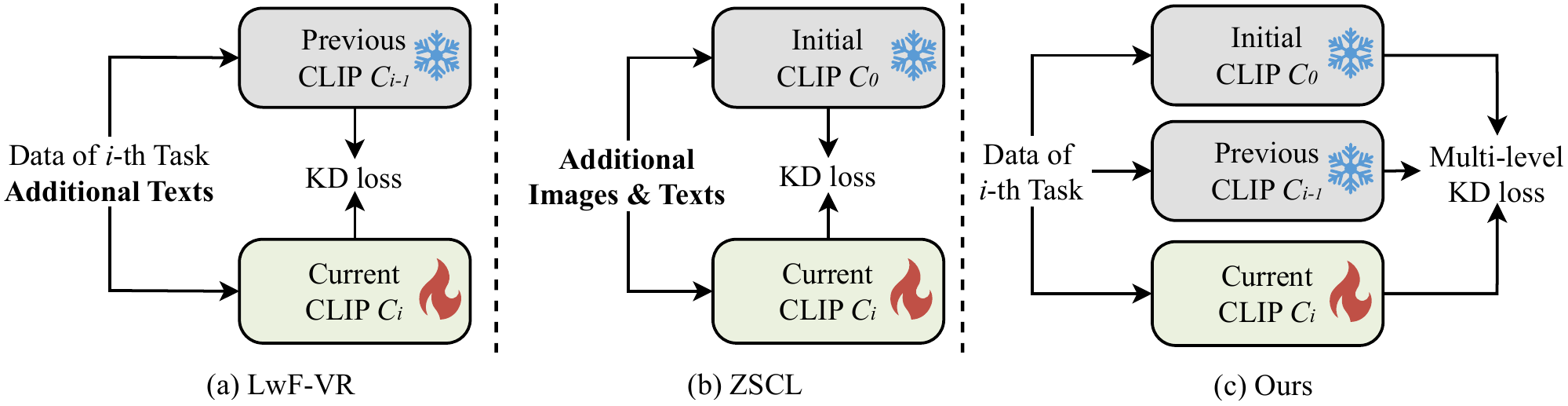}
	\end{center}
	\caption{Illustration of three distillation paradigms and the utilization of additional data, where additional data is marked in bold. (a) LwF-VR\cite{lwfvr} utilizes texts from past tasks (including texts in pretrain) to implement single-teacher distillation between current and previous models. (b) ZSCL\cite{zscl} utilizes reference images and texts to implement single-teacher distillation between current and the initial models. (c) Our method utilizes only training data of current task, and implementing dual-teacher distillation among current, previous and the initial models.}
	\label{distill}
\end{figure*}
Through concerted research efforts, CL-VLMs has developed various strategies to tackle these challenges\cite{moe, lwfvr, zscl, sprompt}. Among these, distillation-based techniques stand out by eliminating the need for replayed samples and enhancing knowledge transfer across tasks. Notable approaches such as LwF-VR \cite{lwfvr} and ZSCL \cite{zscl} have achieved significant milestones. However, these methods still exhibit severe limitations that are pervasive in current distillation-based CL-VLM approaches, thereby hindering further development.

Firstly, the single-teacher distillation paradigm \cite{lwfvr, zscl} fails to impart comprehensive knowledge to the student model, akin to how a single-subject teacher cannot cultivate the most outstanding students. For example, LwF-VR \cite{lwfvr} employs the preceding model, while ZSCL \cite{zscl} leverages the initial model. Both the pretrained and preceding models address either catastrophic forgetting or generalization forgetting, or partially mitigate these issues. As depicted in Fig.\ref{tu1}(a), single-teacher distillation from the preceding model is less effective in mitigating generalization forgetting, a similar trend is observed with initial model distillation in Fig.\ref{tu1}(b).

Secondly, existing distillation methods inadequately leverage the multimodal information available in the original training dataset. For example, LwF-VR simplistically samples texts from the pretraining dataset for distillation, while ZSCL employs ImageNet \cite{imagenet} as a reference dataset. The extensive reference dataset not only demands substantial memory and computational resources but also contains overly rich semantic information, leading to significant performance degradation in ZSCL when the reference dataset is altered \cite{zscl}. Consequently, both methods overly rely on supplementary data for distillation rather than leveraging the knowledge from the original training set, as illustrated in Fig.\ref{distill}(a) and (b).


In general, the aforementioned deficiencies arise primarily from the imperfect integration and transfer of knowledge between modalities. According to the Platonic Representation Hypothesis \cite{platonic}, knowledge across different modalities exhibit convergent properties and equal importance in the platonic space after adaptation through extensive data. Besides, a single-teacher distillation approach is insufficient for forming a learning process akin to that of human beings. To address these issues, we draw on Knowledge Integration Theory (KIT) \cite{bell1995knowledge}, which is well-known in the field of education. KIT emphasizes constructing new knowledge by establishing cross-modal connections between new information and prior knowledge. It outlines four stages for optimal human education: Eliciting Ideas, Adding New Ideas, Distinguishing Ideas, and Making Connections. Our approach aims to enhance existing distillation-based CL-VLMs by adhering to these four stages.

Guided by KIT, we propose MulKI, a Multi-Stage Knowledge Integration network designed to emulate the human learning process in distillation methods. MulKI achieves this through four stages. First, in the Eliciting Ideas stage, we utilize prototypes to align different modalities, identifying and aligning core knowledge from each modality. Next, for Adding New Ideas stage, MulKI enriches existing knowledge by constructing fine-grained intra- and inter-modality relationships with prototypes, introducing new knowledge and information of teacher models from three levels. As for the Distinguishing Ideas stage, we refine and combine the knowledge from two teacher models through adaptively adjusting the sample-wise learning focus with respect to different teachers. Finally, in the Making Connections stage, MulKI integrates knowledge between models,  balancing preceding and new knowledge from both intra- and inter-task perspectives. By following these four stages, MulKI not only facilitates comprehensive cross-modal knowledge integration, but also enhances the model's adaptability in diverse data environments. Unlike LwF-VR\cite{lwfvr} and ZSCL\cite{zscl}, MulKI requires no additional data or preservation of prototypes from prior tasks. Prototypes are promptly purged from memory upon task completion, requiring no data storage beyond model weights and using only the training dataset and pretrained CLIP weights. Figure \ref{distill}(c) shows the paradigm of our method.

In summary, our contributions are summarized as follows:\par
(1) We introduce the Multi-Stage Knowledge Integration (MulKI) network, inspired by the KIT educational theory. Mimicking the human learning process, MulKI enhances existing distillation-based methods, fostering outstanding student model.

(2) We design a four-stage learning process for MulKI akin to that of human beings, including Eliciting Ideas, Adding New Ideas, Distinguishing Ideas, and Making Connections. MulKI constructs high-quality knowledge by establishing cross-modal connections between new information and prior knowledge. 

(3) Our MulKI explores internal relationships across modalities by leveraging prototypes. It integrates new knowledge from various perspectives under a dual-teacher paradigm, without the need for extra data. The learning focus of the two teacher models is adaptively adjusted for optimal performance.




\section{Related work}
\subsection{Continual Learning}
Continual learning aims at acquiring knowledge continually across a series of tasks without forgetting old knowledge\cite{lwf}. Traditional CL approaches can be broadly categorized into four types: replay-based methods, regularization-based methods, structure-based methods and PEFT-based methods.  \par
Replay-based methods explicitly\cite{gdumb, latent_replay, icarl, der} or implicitly \cite{dgr, lgm, 2018pseudo} retain a small number of exemplars from previous tasks to participate in subsequent tasks. Exemplars stored explicitly are typically original samples but can also be features or output distributions. Exemplars stored implicitly are typically pseudo-samples or features generated by GANs\cite{gan} or VAEs\cite{vae}. 
Regularization-based methods introduce penalty terms on parameters\cite{Memoryawaresynapses, ewc, lee2017overcoming} or responses\cite{dhar2019learning, podnet, lucir}, to preserve old knowledge, mitigating catastrophic forgetting at the expense of model's plasticity. The pioneering method in this category is EWC\cite{ewc}, which constrains the updates of parameters that are vital to old knowledge. A popular branch in this group is to explore the distillation methods\cite{ckdf, lijin, lu2024pamk}, which encourage student models to maintain consistent responses with teacher models to retain knowledge. 
Structure-based methods introduce new branches for new tasks\cite{hu2023dense, ye2023self, der2021}. For instance, Dytox\cite{dytox} and DEN\cite{yoon2017lifelong} add new parameters for new tasks. \cite{expert} introduces new structures for new tasks and utilize gating mechanisms to select task-specific and task-shared structures.\par

With the appearance of pretrained ViTs\cite{vit}, PEFT-based methods\cite{l2p, dualprompt, coda} emerge as a new category of continual learning methods. These methods keep the backbone frozen and only finetune lightweight learnable parameters to adapt to downstream tasks\cite{l2p, lora, zhao2024learning}. For instance, CODA-prompt\cite{coda} finetunes a set of prompt components for all tasks. DualPrompt\cite{dualprompt} adds new prompts for new tasks.\par 
Our proposed MulKI mainly utilizes distillation and falls into distillation methods. It is particularly designed for VLMs.

\begin{figure*}
	\begin{center}
		\includegraphics[scale=1.5]{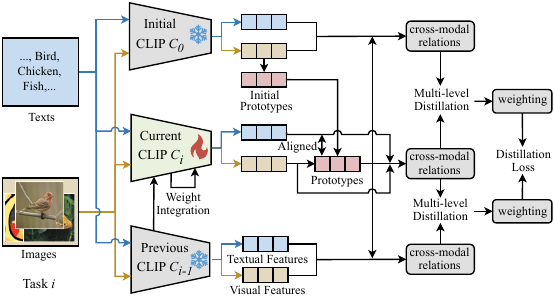}
	\end{center}
	\caption{Overview of the proposed MulKI network. It utilizes only images and texts from current task to implement dual-teacher distillation. Prototypes are updated with current image features. Then multi-level knowledge distillation is employed on cross-modal relations, which is acquired under the guidance of prototypes. The final multi-level distillation loss is computed by combining the loss from two teacher models. }
	\label{datu}
\end{figure*}

\subsection{Continual Learning on VLMs}
There are already several notable works on CL-VLMs, which can be roughly categorized into structure-based methods\cite{moe, proof}, PEFT-based methods~\cite{sprompt,coleclip}, and distillation-based methods~\cite{zscl,lwfvr}. 
MoE\cite{moe} is a typical structure-based methods, which inserts mixture-of-expert structures as MoE-adapters into CLIP and adds additional experts blocks for each new task. Besides, MoE utilizes additional dataset to train Distribution Discriminative Auto-Selector, which serves as task-label selector. 
PEFT-based methods finetune lightweight learnable parameters for both visual and textual modalities. For example,  S-Prompt\cite{sprompt} explores domain-incremental learning of CLIP by utilizing image-text prompts pool. It incrementally adds light weight prompts for distinct domain and retrieve the prompts by performing K-means and K-NN.\par
Recently, distillation-based methods have received increasing attentions. For example, LwF-VR\cite{lwfvr} stands out as one of the earliest distillation-based methods, utilizing replayed vocabularies and model from the previous task for single-teacher distillation. ZSCL\cite{zscl} leverages a vast reference dataset containing rich semantic information and pretrained CLIP model for single-teacher distillation. The large reference dataset ZSCL relies in distillation maintains the feature space as unchanged as possible. 
In summary, currently distillation-based approaches play an important role in CL-VLMs, and most existing methods more or less require additional data. \par

In contrast, our proposed MulKI network employs prototypes to conduct multi-stage distillation within a dual-teacher framework. Drawing from human learning process, this approach effectively integrates cross-modal information from the dataset without the need for additional data, surpassing the capabilities of existing methods.

\section{Methodology}
\subsection{Task Definition}
We choose Contrastive Language-Image Pre-train (CLIP)\cite{clip} model to implement CL-VLMs. In CL-VLMs, tasks are associated with distinct datasets, denoted as $\left\{ \mathcal{T}_1,\mathcal{T}\!_2,...,\mathcal{T}\!\:_{\!\:\!\!\!N} \right\}$, then the CLIP model is finetuned on each task sequentially. $\mathcal{T}_i=\left\{X^i,Y^i,T^i \right\}$ represents the data of the $\textit{i}^{th}$ task, where $X^i$ is input images, and $Y^i$ is the corresponding labels, and $T^i$ maps $Y^i$ into object names with predefined template. In addition to alleviating catastrophic forgetting, another key objective in CL-VLMs is to maintain the inherent zero-shot generalization capability of CLIP model. The model fine-tuned on task $\mathcal{T}_i$ is denoted as $C_i$, which is iteratively built upon the previous model $C_{i-1}$. The initial CLIP model is denoted as $C_0$. \par
\subsection{MulKI Method}
\textit{\textbf{1) Framework Overview }}\par
The proposed MulKI aims at achieving continual learning in vision-language models by emulating the human learning process. The framework is built upon the CLIP model, and the overall structure is illustrated in Fig. \ref{datu}. \par
We leverage prototypes for constructing knowledge in MulKI, the initial prototypes are created by averaging the visual features extracted with $C_0$, which are then updated per iteration. We then employ contrastive learning to align across modalities. Based on this, we construct high-quality knowledge with prototypes by employing multi-level dual-teacher knowledge distillation. Furthermore, we employ a distribution similarity based strategy to adaptively distinguish and balance the knowledge derived from the two teacher models. Additionally, we establish connections between models, integrating preceding and new knowledge. \par
\textit{\textbf{2) Initialization and Update of Prototypes }}\par
Before the training of each task, the original CLIP model is utilized to compute the visual features of all samples from each task, then the initial prototypes are computed by averaging these features for each class, take class \textit{c} as example, the corresponding initial prototype is constructed as follows:
\begin{equation}
p_{init}^{c}=Norm\left( \frac{1}{n}\sum\limits_{k=1}^n{C_0\left( x_{k}^{c} \right)} \right),
\end{equation}
where \textit{c} denotes the class index, $x_{i}^{c}$ represents the image in class \textit{c}, \textit{n} represents the image numbers of class \textit{c}, and \textit{Norm} represents the normalization function. For brevity, in the above equations, $C_0\left( x \right)$ denotes extracting the feature of sample \textit{x} according to model $C_0$, the same below. \par
Due to the substantial computational resources required for directly updating prototypes with the entire dataset during training, we adopt a sliding average update strategy. During the training process of each task, prototypes are updated iteratively. For class \textit{c}, the prototype is updated as follows:
\begin{equation}
p_c=Norm\left( \gamma* p_c+\left( 1-\gamma \right) \frac{1}{n}\sum\limits_{k=1}^n{C_i\left( x_{k}^{c} \right)} \right) ,\label{pro_cons}
\end{equation}
where the first term represents the prototypes from the previous iteration, while the second represents the new prototypes acquired in current iteration. $\gamma$ is a sliding parameter, which controls the update speed of prototypes. In practice, the value of $\gamma$ gradually increases by iteration, i.e. $\gamma=\gamma+\textit{step}$, and eventually stabilizes at a large value. 
 
\textit{\textbf{3) Cross-modal Symmetric Alignment}}\par
To eliciting core knowledge from both modalities, we ensure alignment between prototypes and text by utilizing a contrastive learning loss, which is in consistent with that is used in CLIP pre-training. The core idea of alignment is to maximize the similarity between prototypes and text embeddings of the same class by optimizing a symmetric cross-entropy loss over the similarity score matrices. The prototype-to-text loss $\mathcal{L}_{p2t}$ and text-to-prototype loss $\mathcal{L}_{t2p}$ are formulated as:
\begin{equation}
\mathcal{L}_{p2t}=-\frac{1}{C}\sum_{k=1}^C{\log \frac{\exp \left( sim\left( P_k^m,T_k^m \right) /\tau \right)}{\sum_{j=1}^C{\exp \left( sim\left( P_k^m,T_j^m \right) /\tau \right)}}}, 
\end{equation}
\begin{equation}
\mathcal{L}_{t2p}=-\frac{1}{C}\sum_{k=1}^C{\log \frac{\exp \left( sim\left( P_k^m,T_k^m \right) /\tau \right)}{\sum_{j=1}^C{\exp \left( sim\left( P_j^m,T_k^m \right) /\tau \right)}}},
\end{equation}
where \( C \) represents the total number of classes in current task, \textit{sim} denotes the cosine similarity function, \( P_k^m \) and \( T_k^m \) denote the \( k \)-th prototype and text embedding from the $m^{th}$ task respectively, and $\tau$ stands for the temperature parameter. \par
Therefore, combining $\mathcal{L}_{p2t}$ and  $\mathcal{L}_{t2p}$, the cross-modal symmetric alignment loss is defined as:
\begin{equation}
\mathcal{L}_{CSA}=\left( \mathcal{L}_{p2t}+\mathcal{L}_{t2p} \right) /2.
\end{equation}

\textit{\textbf{4) Multi-level Dual-teacher Knowledge Distillation}}\par

To construct and add high-quality new knowledge to the student model, it is crucial to fully leverage the multimodal information in the training dataset and ensure comprehensive knowledge transfer by using two teacher models. We adopt dual-teacher distillation and propose a multi-level knowledge distillation method to preserve knowledge from both teacher models thoroughly.

We use $C_0$ for distillation to retain the zero-shot generalization capability of the initial CLIP and $C_{i-1}$ to preserve accumulated knowledge, thus alleviating both generalization forgetting and catastrophic forgetting during training. To achieve effective distillation without additional data, we comprehensively distill the teacher models' knowledge through multi-level distillation, including feature distillation, intra-modal relationship distillation, and inter-modal distribution distillation.\par
Firstly, we employ feature distillation on visual feature instances, aiming to capture coarse-grained distribution of the teacher model in a simple yet efficient manner. We use a mean squared loss to implement effective feature distillation, minimizing the mean squared error between the student and teacher models' feature outputs. This ensures the student model's outputs remain consistent with the teacher model's outputs. Taking the feature distillation between $C_0$ and $C_i$ as an example, the Feature Distillation loss is as follows:

\begin{equation}
\overline{\mathcal{L}_{FD}}=\frac{1}{\left| X^m \right|}\sum_{x\in X^m}{\lVert C_0\left( x \right) -C_i\left( x \right) \rVert _{2}^{2}},
\end{equation}
where $X^m$ denotes the image inputs of the $m^{th}$ task. \par

Secondly, intra-modal relationship distillation captures the similarity relationships between visual feature instances and prototypes within a task, ensuring the student model inherits these relationships from the teacher model during fine-tuning. We calculate the similarity between all visual feature instances and prototypes for the current task using both models. We then minimize the Frobenius norm of the difference between the relation matrices from the two models to maintain consistency in the instance-prototype similarity relationships. When distilling from model $C_0$, the Intra-modal Relationship Distillation loss is formulated as:
\begin{equation}
\resizebox{0.9\linewidth}{!}{$
\overline{\mathcal{L}_{IRD}}=\mathbb{E}_{X^m\sim \mathcal{T}_t}\lVert sim\left( C_0\left( X^m \right) ,P^m \right) -sim\left( C_i\left( X^m \right) ,P^m \right) \rVert _F
$},
\end{equation}
where $P^m$ denotes the prototypes of the $m^{th}$ task, \textit{sim} denotes the similarity function, and $\lVert \cdot \rVert _F$ represents the Frobenius norm. \par

Finally, the inter-modal distribution distillation emphasizes the preservation of cross-modal distribution information including image-text distribution and prototype-text distribution, ensuring that student model retains the distribution of teacher model while learning new distribution. We compute these distributions using both teacher and student models and constrain the student's distribution to match the teacher's. This is achieved by employing cross-entropy loss for distillation. When distilling from model $C_0$, the image-text distillation loss is defined as follows:
\begin{equation}
\overline{\mathcal{L}_{i2t}}=-\frac{1}{\left| X^m \right|}\sum_{x\in X^m}{\mathcal{P}_{C_0}\left( x,T^m \right) \log \mathcal{P}_{C_i}\left( x,T^m \right)},\label{i2t}
\end{equation}
where $X^m$ and $T^m$ denotes the images and texts input of $m^{th}$ task, and $\mathcal{P}_{C_i}\left( x,T^m \right) $ is formulated as:
\begin{equation}
\mathcal{P}_{C_i}\left(x,T^m \right)=\frac{\exp \left( sim\left( C_i\left( x \right) ,C_i\left( T^m \right) \right) /\tau \right)}{\sum\nolimits_{t\in T^m}{\exp \left( sim\left( C_i\left( x \right) ,C_i\left( t \right) \right) /\tau \right)}}.\label{p}
\end{equation}
\par
Then prototype-text distillation loss when distilling from $C_0$ is formulated as:
\begin{equation}
\begin{split}
\overline{\mathcal{L}_{p\&t}}=\label{lpt}
-\frac{1}{C}\sum_{k=1}^{C}{\mathcal{P}_{C_0}\left(P_{k}^{m},T^m \right) \log \mathcal{P}_{C_i}\left(P_{k}^{m},T^m \right)}+\\
-\frac{1}{C}\sum_{k=1}^{C}{\mathcal{P}_{C_0}\left(T_{k}^{m},P^m \right) \log \mathcal{P}_{C_i}\left(T_{k}^{m},P^m \right)},
\end{split}
\end{equation}
in which function $\mathcal{P}$ is defined to be consistent with formula (\ref{p}). Thus the Inter-modal Distribution Distillation loss from model $C_0$ is acquired with formulas (\ref{i2t}) and (\ref{lpt}):
\begin{equation}
\overline{\mathcal{L}_{IDD}}=\overline{\mathcal{L}_{i2t}}+\overline{\mathcal{L}_{p\&t}}.
\end{equation}
\textit{\textbf{5) Similarity-based Weighting Strategy}}\par

To effectively distinguish and combine knowledge from two teacher models, it is infeasible to assign equal weights to their losses, despite their equal importance. Due to dataset diversity and sample variability, some samples perform well with model $C_0$, mitigating generalization forgetting, while others perform otherwise. Assigning equal weights may lead to a suboptimal model as it fails to accommodate the specific circumstances of individual samples. \par 

To address this, we propose a distribution similarity-based weighting strategy. This strategy assigns a higher weight to the model whose output distribution differs more from the current model, dynamically adjusting the learning focus for each sample. This enables the student model to better integrate the advantages of both teacher models, achieving more effective and balanced learning outcomes. Specifically, for a given sample $\textit{x}$, when distilling from model $C_0$, the weight for the loss is computed as follows:
\begin{equation}
r_0\left( x \right) =\frac{1/\exp \left( s_0\left( x \right) \right)}{1/\exp \left( s_0\left( x \right) \right) +1/\exp \left( s_{i-1}\left( x \right) \right)},
\end{equation}
where $s_0$ is defined as:
\begin{equation}
\begin{aligned}
s_0\left( x \right) &=sim\left( \mathcal{P}_{C_0}\left( x,T^m \right) ,\mathcal{P}_{C_i}\left( x,T^m \right) \right).\\
\end{aligned}
\end{equation}
Then $r_0$ and $s_{i-1}$ are defined alike. This strategy is applied to all distillation losses except for $\mathcal{L}_{p\&t}$ since this strategy is designed to adapt the diversity of individual samples, whereas the prototypes in $\mathcal{L}_{p\&t}$ represent the overall attribute of the corresponding class.\par
Consequently, the overall Multi-level Dual-teacher Knowledge Distillation loss for model $C_0$ with similarity-based weight strategy is formulated as:
\begin{equation}
\overline{\mathcal{L}_{MDD}}=r_0(\overline{\mathcal{L}_{FD}}+\alpha\overline{\mathcal{L}_{IRD}}+\beta\overline{\mathcal{L}_{IDD}}),
\end{equation}
where $\alpha$ and $\beta$ are hyper-parameters, and $r_0$ above is replaced with 0.5 if this strategy is not adopted. Note that $r_0$ does not apply to the loss $\overline{\mathcal{L}_{p\&t}}$ in $\overline{\mathcal{L}_{IDD}}$.  \par
\textit{\textbf{6) Weight Integration in Parameter Space}}\par
Finally, we establish connections with respect to the weight of models from both inter- and intra-task perspectives, integrating preceding and new knowledge. Since the knowledge acquired by the model during training is encapsulated within its parameters, we achieve knowledge retention by employing regularization techniques. Here, we employ Weight Consolidation (WC)\cite{ewc} as well as Weight Ensemble (WE)\cite{zscl} to regularize the weight parameters from inter-task and intra-task perspectives respectively.\par
Denote the model weights at task \( t \) as \( \theta_t \), WC applies a regularization loss on weight parameters between the current model and the previous one:
\begin{equation}\label{wc}
\mathcal{L}_{WC} = \sum_i (\theta_{t,i} - \theta_{t-1,i})^2 \   
\end{equation}
where \( \theta_{t,i} \) represents the \( i \)-th weight parameter at task \( t \), and \( \theta_{t-1,i} \) represents the corresponding weight parameter at the previous task. WC helps to preserve the knowledge acquired from previous tasks by constraining the updates of the weight parameters.\par
WE\cite{zscl}, on the other hand, performs a weighted averaging on weight parameters from intra-task perspective. Denote the model of task \textit{t} after \( k \) iterations as \( \theta_{t,k} \), the weighted averaging on parameters is computed as follows:
\begin{equation}
\hat{\theta}_{t,k,i}=\left\{ \begin{array}{l}\label{we}
	\theta _{t-1,i}\ \ \ \ \ \ \ \ \ \ \ \ \ \ \ \ \ \ \ \ \ \ \ \ k=0\\
	\frac{1}{m+1}\theta _{t,k,i}+\frac{m}{m+1}\hat{\theta}_{t,k-1,i}\ \ every\ I\ iters\\
\end{array}, \right.   
\end{equation}
where $\hat{\theta}_{t,k,i}$ represents the model after the \( m \)-th weighted averaging, in which $m=k/I$. This approach helps to mitigate forgetting by periodically averaging the weights, which ensures the model retains knowledge in the early stages of training.\par
Furthermore, in order to attach more importance on maintaining the model’s stability, we modify WE into Extended Weight Ensemble (EWE). During training time, instead of always keeping $\hat{\theta}_{t,k}$ frozen and making it averaged with $\theta _{t,k}$, we replace $\theta _{t,k}$ with the averaged $\hat{\theta}_{t,k}$ for every $\eta$ Weight Ensemble operations (average operations) as in formula (\ref{we}). The replacement is formulated as:
\begin{equation}
\theta _{t,k,i}=\hat{\theta}_{t,k,i}\ ,\ \ every\ \eta I\ iters.\label{EWE}
\end{equation}
Consequently, the EWE strategy is comprised with formulas (\ref{we}) and (\ref{EWE}).

\subsection{The Overall Objective }
The overall objective function, which consists of cross-modal symmetric alignment loss and multilevel knowledge distillation loss, is as follows:
\begin{equation}
\mathcal{L}_{MulKI}=\mathcal{L}_{ce}+\lambda_1 \mathcal{L}_{CSA}+\lambda_2 \mathcal{L}_{MDD},
\end{equation}
where $\mathcal{L}_{ce}$ denotes the classification loss, $\lambda_1$ and $\lambda_2 $ are hyper-parameters, and $\mathcal{L}_{MDD}$ consists of loss from both teacher models,
\begin{equation}
\mathcal{L}_{MDD}=\overline{\mathcal{L}_{MDD}}+\widehat{\mathcal{L}_{MDD}},
\end{equation}
where $\widehat{\mathcal{L}_{MDD}}$ denotes the distillation loss from the model $C_{i-1}$. Additionally, when WC is adopted, $\mathcal{L}_{WC}$ in formula (\ref{wc}) in added into the above objective function.\par

\section{Experiments}
\subsection{Datasets} 
Three popular datasets are selected in our image classification experiments: CIFAR-100\cite{cifar}, TinyImageNet\cite{der2021} and MTIL\cite{zscl}. \par
CIFAR-100\cite{cifar} consists of 100 classes, each class has 600 samples of 32$\times$32 color images which are divided into 500 for training and 100 for testing, including 5000 training samples and 1000 test samples. TinyImageNet\cite{der2021} is a subset of ImageNet \cite{imagenet}, which contains 200 classes, and each class has 500 samples with 64$\times$64 color images for training. For CIFAR-100, We split it evenly into 10, 20 and 50 sequential tasks, each of which includes 10, 5 and 2 classes. While for TinyImageNet, we first arrange base training including 100 classes, then the remaining 100 classes are evenly split into 5, 10 and 20 sequential tasks, each of which includes 40, 20 and 10 classes. We conduct Class Incremental Learning experiments on these two datasets.\par
\begin{table}[t]
    \caption{Performance (\%) Comparison of state-of-the-art CL methods on \textbf{CIFAR-100} benchmark in class-incremental setting}\label{cifar_cil}
        \centering
	\begin{tabular}{lcc|cc|cc}
            \toprule 
            \multirow{2}{*}{Method} & \multicolumn{2}{c}{10 step} & \multicolumn{2}{c}{20 step} & \multicolumn{2}{c}{50 step} \\
            \cline{2-7} 
            \noalign{\smallskip}
			 & 	Avg. & Last &Avg. & Last &Avg. &Last  \\
		    \midrule
            LUCIR \cite{lucir}  & 58.66 & 43.39& 58.17 & 40.63 & 56.86 & 37.09 \\
            BiC\cite{bic} &68.80 &53.54 &66.48&47.02 & 62.09 & 41.04\\
            RPSNet\cite{rspnet} &68.60&57.05 	&-	&-	&-	&- \\
		  PODNet\cite{podnet} &58.03&41.05&53.97&35.02&51.19&32.99\\
            DER\cite{der2021}  & 74.64 & 64.35 & 73.98 & 62.55 & 72.05 & 59.76\\
            DyTox++\cite{dytox} & 74.10 & 62.34 & 71.62 & 57.43 & 68.90 & 51.09 \\
            \midrule
            CLIP Zero-shot\cite{clip} &74.47  & 65.92 & 75.20 & 65.74  & 75.67 & 65.94 \\
            Continual FT & 65.46& 53.23& 59.69& 43.13&39.23 &18.89\\
            LwF\cite{lwf} &65.86 &48.04 &60.64 &40.56 &47.69 &32.90\\
            iCaRL\cite{icarl} & 79.35 &70.97&73.32 &64.55 & 71.28&59.07\\
            LwF-VR\cite{lwfvr} &78.81 &70.75 &74.54 &63.54 &71.02 & 59.45\\
            ZSCL\cite{zscl} &82.15&73.65&80.39&69.58&79.92&67.36\\
            MoE\cite{moe} &\underline{85.21} &\underline{77.52} &\underline{83.72} &\underline{76.20} 	&\underline{83.60} 	&\underline{75.24} \\
		  \cellcolor{gray!30}\textbf{MulKI(Ours)} \cellcolor{gray!30}&\textbf{87.17}\cellcolor{gray!30}&\textbf{80.74} \cellcolor{gray!30}&\textbf{85.75}\cellcolor{gray!30}&\textbf{78.50} \cellcolor{gray!30}&\textbf{84.93}\cellcolor{gray!30}&\textbf{75.47}\cellcolor{gray!30} \\
		\bottomrule
	\end{tabular}
\end{table}
\begin{table}[t]
    \caption{Performance (\%) Comparison of state-of-the-art CL methods on \textbf{TinyImageNet} benchmark in class-incremental setting}\label{tinyimagenet_cil}
        \centering
	\begin{tabular}{lcc|cc|cc}
            \toprule 
            \multirow{2}{*}{Method} & \multicolumn{2}{c}{5 step} & \multicolumn{2}{c}{10 step} & \multicolumn{2}{c}{20 step} \\
            \cline{2-7} 
            \noalign{\smallskip}
			 & 	Avg. & Last &Avg. & Last &Avg. &Last  \\
		    \midrule
            EWC\cite{ewc} & 19.01 & 6.00&15.82 &3.79 & 12.35&4.73  \\
            EEIL\cite{eeil} &47.17 & 35.12&45.03 &34.64 & 40.41&29.72\\
            LUCIR\cite{lucir} &50.30 & 39.42& 48.58& 37.29& 42.84& 30.85  \\
		    MUC\cite{muc} &32.23&19.20 & 26.67& 15.33& 21.89&10.32  \\
            PASS\cite{pass} &49.54 & 41.64&47.19& 39.27& 42.01& 32.93 \\
            DyTox\cite{dytox} & 55.58& 47.23&52.26 & 42.79& 46.18&36.21   \\
            \midrule
            CLIP Zero-shot\cite{clip} & 69.62& 65.30& 69.55& 65.59& 69.49&65.30  \\
            Continual FT &61.54 &46.66 &57.05 & 41.54& 54.62 & 44.55  \\
            LwF\cite{lwf} &60.97 &48.77 &57.60 &44.00 &54.79 &42.26  \\
            iCaRL\cite{icarl}  & 77.02& 70.39& 73.48&65.97 &69.65 &64.68  \\
            LwF-VR\cite{lwfvr}& 77.56&70.89 &74.12 &67.05 & 69.94&63.89  \\
            ZSCL\cite{zscl} &80.27&73.57&78.61&71.62& 77.18&68.30  \\
MoE\cite{moe}&\underline{81.12}&\underline{76.81}&\underline{80.23}&\underline{76.35}&\underline{79.96}&\underline{75.77}   \\		    
            \cellcolor{gray!30}\textbf{MulKI(Ours)} \cellcolor{gray!30}&\textbf{82.54}\cellcolor{gray!30}&\textbf{78.36} \cellcolor{gray!30}&\textbf{82.28}\cellcolor{gray!30}&\textbf{77.56} \cellcolor{gray!30}&\textbf{81.77}\cellcolor{gray!30}&\textbf{76.65}\cellcolor{gray!30}  \\
		\bottomrule
	\end{tabular}
\end{table}
\begin{table*}
    \caption{Comparison with state-of-the-art methods on MTIL benchmark in terms of ``Transfer'', ``Avg.'', and ``Last'' scores (\%). We label the best and second methods with \textbf{bold} and \underline{underline} styles. The top block indicates the direct adaptions of CLIP on each task}\label{mtil_order1}
    \centering
    \resizebox{1.0\linewidth}{!}{
    \begin{tabular}{c>{\raggedright\arraybackslash}p{3cm} >{\centering\arraybackslash}p{1cm} >{\centering\arraybackslash}p{1cm}>{\centering\arraybackslash}p{1cm} >{\centering\arraybackslash}p{1cm} >{\centering\arraybackslash}p{1cm}>{\centering\arraybackslash}p{1cm} >{\centering\arraybackslash}p{1cm} >{\centering\arraybackslash}p{1cm}>{\centering\arraybackslash}p{1cm} >{\centering\arraybackslash}p{1cm} >{\centering\arraybackslash}p{1cm} |>{\centering\arraybackslash}p{1.5cm}}
        \toprule
           & {\hspace{1em}} Method & \makecell[c]{\rotatebox{90}{Aircraft~}} & \makecell[c]{\rotatebox{90}{Caltech101~}} & \makecell[c]{\rotatebox{90}{CIFAR100~}} & \makecell[c]{\rotatebox{90}{DTD~}} & \makecell[c]{\rotatebox{90}{EuroSAT~}} & \makecell[c]{\rotatebox{90}{Flowers~}} & \makecell[c]{\rotatebox{90}{Food~}} & \makecell[c]{\rotatebox{90}{MNIST~}} & \makecell[c]{\rotatebox{90}{OxfordPet~}} & \makecell[c]{\rotatebox{90}{Cars~}} & \makecell[c]{\rotatebox{90}{SUN397~}} & \makecell[c]{{\textit{Average}}} \\

        \midrule
        
            \multirow{2}{*}{\rotatebox{90}{CLIP}}& {\hspace{1em}}Zero-shot\cite{clip} & 24.3 & 88.4 & 68.2 & 44.6 & 54.9 & 71.0 & 88.5 & 59.4 & 89.0 & 64.7 & 65.2 & 65.3  \\
            & {\hspace{1em}}Full Fine-tune & 62.0 & 95.1 & 89.6 & 79.5 & 98.9 & 97.5 & 92.7 & 99.6 & 94.7 & 89.6 & 81.8 & 89.2  \\
            
            \midrule\midrule
            \multirow{8}{*}{\rotatebox{90}{\textbf{Transfer}}} &{\hspace{1em}}Continual-FT & -& 67.1 & 46.0 & 32.1 & 35.6 & 35.0 & 57.7 & 44.1 & 60.8 & 20.5 & 46.6 & 44.6 \\
            & {\hspace{1em}}LwF\cite{lwf}  &  -& 74.5 & 56.9 & 39.1 & \underline{51.1} & 52.6 & 72.8 & 60.6 & 75.1 & 30.3 & 55.9 & 58.9 \\
            & {\hspace{1em}}iCaRL\cite{icarl}  &  -& 56.6 & 44.6 & 32.7 & 39.3 & 46.6 & 68.0 & 46.0 & 77.4 & 31.9 & 60.5 & 50.4 \\
            & {\hspace{1em}}LwF-VR\cite{lwfvr}  & -& 77.1 & 61.0 & 40.5 & 45.3 & 54.4 & 74.6 & 47.9 & 76.7 & 36.3 & 58.6 & 57.2  \\
            & {\hspace{1em}}WiSE-FT\cite{wiseft}  & -& 73.5 & 55.6 & 35.6 & 41.5 & 47.0 & 68.3 & 53.9 & 69.3 & 26.8 & 51.9 & 52.3  \\
            & {\hspace{1em}}ZSCL\cite{zscl}  &- & 86.0 & 67.4 & \underline{45.4} & 50.4 & 69.1 & 87.6 & \underline{61.8} & 86.8 & 60.1 & \underline{66.8} & 68.1 \\
            & {\hspace{1em}}MoE\cite{moe} &-&\textbf{87.9} & \underline{68.2} & 44.4 &49.9 &\underline{70.7} & \textbf{88.7} &59.7 &\underline{89.1} &\textbf{64.5} &65.5 &\underline{68.9}\\
            
            \cellcolor{gray!30}& {\hspace{1em}}\cellcolor{gray!30}\textbf{MulKI(ours)} \cellcolor{gray!30}&-\cellcolor{gray!30}&\underline{87.8}\cellcolor{gray!30}&\textbf{69.0}\cellcolor{gray!30}\cellcolor{gray!30}&\textbf{46.7}\cellcolor{gray!30}&\textbf{51.8}\cellcolor{gray!30}&\textbf{71.3}\cellcolor{gray!30}&\underline{88.3}\cellcolor{gray!30}&\textbf{64.7}\cellcolor{gray!30}&\textbf{89.7}\cellcolor{gray!30}&\underline{63.4}\cellcolor{gray!30}&\textbf{68.1}\cellcolor{gray!30}&\textbf{70.1}\cellcolor{gray!30} \\
            \midrule 
            \multirow{8}{*}{\rotatebox{90}{\textbf{Average}}} &{\hspace{1em}}Continual-FT    & 25.5 & 81.5 & 59.1 & 53.2 & 64.7 & 51.8 & 63.2 & 64.3 & 69.7 & 31.8 & 49.7 & 55.9 \\
            & {\hspace{1em}}LwF\cite{lwf}  & 36.3 & 86.9 & 72.0 & 59.0 & 73.7 & 60.0 & 73.6 & 74.8 & 80.0 & 37.3 & 58.1 & 64.7 \\
            & {\hspace{1em}}iCaRL\cite{icarl}  & 35.5 & 89.2 & 72.2 & 60.6 & 68.8 & 70.0 & 78.2 & 62.3 & 81.8 & 41.2 & 62.5 & 65.7 \\
            & {\hspace{1em}}LwF-VR\cite{lwfvr}  & 29.6 & 87.7 & 74.4 & 59.5 & 72.4 & 63.6 & 77.0 & 66.7 & 81.2 & 43.7 & 60.7 & 65.1  \\
            & {\hspace{1em}}WiSE-FT\cite{wiseft} & 26.7 & 86.5 & 64.3 & 57.1 & 65.7 & 58.7 & 71.1 & 70.5 & 75.8 & 36.9 & 54.6 & 60.7  \\
            & {\hspace{1em}}ZSCL\cite{zscl}  & 45.1 & \underline{92.0} & \underline{80.1} & 64.3 & \underline{79.5} & 81.6 & \textbf{89.6} & \underline{75.2} & 88.9 & 64.7 & \underline{68.0} &75.4 \\
            & {\hspace{1em}}MoE\cite{moe} &\underline{50.2}&91.9&\textbf{83.1}& \textbf{69.4}&78.9&\textbf{84.0}&\underline{89.1}&73.7&\underline{89.3}&\textbf{67.7}&66.9&\underline{76.7}\\
            \cellcolor{gray!30}& {\hspace{1em}}\cellcolor{gray!30}\textbf{MulKI(ours)} \cellcolor{gray!30}&\textbf{52.5} 	\cellcolor{gray!30}&\textbf{93.6} 	\cellcolor{gray!30}&79.4 	\cellcolor{gray!30}&\underline{67.0} \cellcolor{gray!30}&	\textbf{79.8} \cellcolor{gray!30}&	\underline{83.9} 	\cellcolor{gray!30}&\textbf{89.6}\cellcolor{gray!30} &	\textbf{77.1} \cellcolor{gray!30}&	\textbf{91.2}\cellcolor{gray!30} &	\underline{67.1} \cellcolor{gray!30}	&\textbf{69.1} \cellcolor{gray!30}	&\textbf{77.3}\cellcolor{gray!30} \\
            \midrule
            \multirow{8}{*}{\rotatebox{90}{\textbf{Last}}} &{\hspace{1em}}Continual-FT & 31.0 & 89.3 & 65.8 & 67.3 & 88.9 & 71.1 & 85.6 & \textbf{99.6} & 92.9 & 77.3 & \underline{81.1} & 77.3 \\
            & {\hspace{1em}}LwF\cite{lwf}  & 26.3 & 87.5 & 71.9 & 66.6 & 79.9 & 66.9 & 83.8 & \textbf{99.6}& 92.1 & 66.1 & 80.4 & 74.6 \\
            & {\hspace{1em}}iCaRL\cite{icarl}  & 35.8 &\textbf{ 93.0 }& 77.0 & 70.2 & 83.3 & 88.5 & \underline{90.4}& 86.7 &93.2 & 81.2 &\textbf{81.9}& 80.1 \\
            & {\hspace{1em}}LwF-VR\cite{lwfvr}  & 20.5 & 89.8 & 72.3 & 67.6 & 85.5 & 73.8 & 85.7 & \textbf{99.6} & 93.1 & 73.3 & 80.9 & 76.6  \\
            & {\hspace{1em}}WiSE-FT\cite{wiseft}  & 27.2 & 90.8 & 68.0 & 68.9 & 86.9 & 74.0 & 87.6 & \textbf{99.6}& 92.6 & 77.8 & 81.3 & 77.7  \\
            & {\hspace{1em}}ZSCL\cite{zscl}  & 40.6 & \underline{92.2} & 81.3 & 70.5 & 94.8 & 90.5 & \textbf{91.9} & 98.7 & \underline{93.9} & \textbf{85.3}& 80.2 & 83.6 \\
            & {\hspace{1em}}MoE\cite{moe}& \textbf{49.8} &\underline{92.2} &\textbf{86.1} &\textbf{78.1}&\underline{95.7}&\textbf{94.3}&89.5&98.1&89.9&81.6&80.0&\underline{85.0}\\
            \cellcolor{gray!30}& {\hspace{1em}}\cellcolor{gray!30}\textbf{MulKI(ours)} \cellcolor{gray!30}&\underline{49.7} \cellcolor{gray!30}&	93.0 	\cellcolor{gray!30}&\underline{82.8}\cellcolor{gray!30} &	\underline{73.7} \cellcolor{gray!30}&\textbf{96.2} \cellcolor{gray!30}	&\underline{92.3} \cellcolor{gray!30}&	\underline{90.4} \cellcolor{gray!30}&	\underline{99.0} \cellcolor{gray!30}&	\textbf{94.8}\cellcolor{gray!30} &	\underline{85.2} \cellcolor{gray!30}&	78.9 \cellcolor{gray!30}	&\textbf{85.1}\cellcolor{gray!30} \\
           
        \bottomrule
    \end{tabular}}
\end{table*}
MTIL (Multi-domain Task Incremental Learning)\cite{zscl} is the cross-domain version of task incremental learning, where different tasks originate from distinct domains. MTIL comprises of 11 tasks, each representing an independent dataset, with a total of 1201 classes. Training and testing are performed using two orders. Order I is in an alphabetical order: Aircraft\cite{aircraft}, Caltech101\cite{caltech}, CIFAR100\cite{cifar}, DTD\cite{dtd}, EuroSAT\cite{eurosat}, Flowers\cite{flower}, Food\cite{food}, MNIST\cite{mnist}, OxfordPet\cite{oxfordpet}, StanfordCars\cite{cars}, SUN397\cite{sun}, while Order II is in a random order: StanfordCars, Food, MNIST, OxfordPet, Flowers, SUN397, Aircraft, Caltech101, DTD, EuroSAT, CIFAR100. We conduct Multi-domain Task Incremental Learning experiments on MTIL benchmark.

\subsection{Implementation Details}
We employ a pretrained CLIP model with a ViT-B/16\cite{vit} image encoder in our experiments. We train 1000 iterations for each task in MTIL while for sequential CIFAR-100 and TinyImageNet, each task is trained only 1 epoch. For the few-shot setting of MTIL, we train 500 iterations for each task. AdamW\cite{adamw} is utilized as optimizer and batch size is fixed at 32 for all experiments. For the update of prototypes, we set initial $\gamma$ as 0 and increase it by less than 0.1 per iteration until it reaches more than 0.9. Specifically, for CIFAR-100, $\gamma$ is increased by 0.04 per iteration until it equals 0.98. All $\tau$ mentioned in above formulas are set as 2 uniformly. In addition, WE\cite{zscl} is utilized for both Multi-domain Task Incremental Learning and Class Incremental Learning settings, while WC\cite{ewc} is only adopted in the former setting. Particularly, in Class Incremental Learning settings, we adopt WE for TinyImageNet and EWE for CIFAR-100.

\begin{table}[t]
\centering
\caption{Performance (\%) Comparison of of state-of-the-art CL methods on MTIL benchmark in Order II
}\label{mtil_order2}
\resizebox{\columnwidth}{!}{%
\begin{tabular}{@{}lcc|cc|cc@{}}
\toprule
Method & Transfer & $\Delta$ & Avg. & $\Delta$ & Last & $\Delta$  \\ \midrule
 CLIP Zero-shot\cite{clip}  &65.4& 	0.0 	&65.3 	&0.0 	  &65.3 &	0.0  \\
 Continual FT &46.6& 	-18.8& 	56.2& 	-9.1 	&67.4 &	2.1   \\
 \midrule
 LwF\cite{lwf}~ &53.2& 	-12.2& 	62.2& 	-3.1 	&71.9 &	6.6   \\
 iCaRL\cite{icarl}~ &50.9& 	-14.5& 	56.9& 	-8.4 	&71.6 &	6.3  \\
 LwF-VR\cite{lwfvr}~ &53.1& 	-12.3& 	60.6& 	-4.7 	&68.3 &	3.0  \\
 WiSE-FT\cite{wiseft}~&51.0& 	-14.4& 	61.5& 	-3.8 	&72.2 &	6.9   \\
 ZSCL\cite{zscl} ~ &64.2& 	-1.2 &	74.5 &	9.2 	&83.4 &	18.1  \\
 MoE\cite{moe}~ &\underline{64.3}& 	\underline{-1.1} &	\underline{74.7} &	\underline{9.4} 	&\underline{84.1} &	\underline{18.8}   \\
 \cellcolor{gray!30}MulKI(ours)~\cellcolor{gray!30}&\textbf{65.6}\cellcolor{gray!30}&\textbf{0.2}\cellcolor{gray!30}&\textbf{75.0}
                               \cellcolor{gray!30}&\textbf{9.7}\cellcolor{gray!30}&\textbf{84.2}\cellcolor{gray!30}&\textbf{18.9}\cellcolor{gray!30}  \\
\bottomrule
\end{tabular}%
}
\end{table}

\subsection{Evaluation Metrics}
For Multi-domain Task Incremental Learning, following ZSCL\cite{zscl}, we utilize three metrics, namely ``Avg.", ``Last", and ``Transfer" to evaluate our method. ``Avg." measures the average accuracy of the model over all tasks.  ``Transfer” assesses the model’s zero-shot transfer capability over subsequent tasks. ``Last" represents the average performance after the whole continual learning process. The ``Current Avg." metric mentioned in Fig. \ref{tu1} is calculated by considering only seen tasks in metric ``Avg." 

For Class Incremental Learning setting, we do not report ``Transfer" and only adopt ``Avg." and ``Last" metrics.

\subsection{Experiments on Class Incremental Learning Setting}
\textit{1) Competitors:} We first evaluate our methods on class incremental learning. Various previous methods are chosen as competitors, including: (1) LUCIR\cite{lucir}; (2) BiC\cite{bic}; (3) RPSNet\cite{rspnet}; (4)  PoDNet\cite{podnet}; (5) DER\cite{der2021}; (6) DyTox\cite{dytox}; (7) EWC\cite{ewc}; (8) EEIL\cite{eeil}; (9) MUC\cite{muc}; (10) PASS\cite{pass}; (11) LwF\cite{lwf}; (12) iCaRL\cite{icarl}; (13) LwF-VR\cite{lwfvr}; (14) ZSCL\cite{zscl}; (15) MoE\cite{moe}, in which (11)-(15) are based on CLIP or re-implemented with CLIP backbone, the rest cannot be easily adapted due to their special design. \par

In addition, we also report two special competitors, including CLIP Zero-shot and Continual-FT, where the former represents the Zero-shot Transfer capability of CLIP pretrained model without any finetuning, and the latter means that no continual learning strategy is adopted in the continual finetune of CLIP pretrained model. We report ``Avg." and ``Last" metrics in Class Incremental Learning setting.\par
\textit{2) Results:} Table \ref{cifar_cil} and \ref{tinyimagenet_cil} report the results of the comparisons on CIFAR-100 and TinyImageNet. It is observed that our approach outperforms the competitors in all cases. Specifically, for the 10-steps setting of sequential CIFAR-100 dataset, the proposed MulKI outperforms the state-of-the-art competitors at least 1.96\% and 3.22\% for ``Avg." and ``Last" respectively. As learning steps increases, continual learning becomes more challenging, our MulKI still outperforms other methods. Furthermore, MulKI beats all other methods on sequential TinyImageNet dataset.\par
\begin{table}[t]
\centering
\caption{Performance (\%) Comparison of state-of-the-art CL methods on MTIL benchmark in Order I under few-shot setting
}\label{few1}
\begin{tabular}{@{}lcc|cc|cc@{}}
\toprule
Method & Transfer & $\Delta$ & Avg. & $\Delta$ & Last & $\Delta$  \\ \midrule
 CLIP Zero-shot\cite{clip}  & 69.4 &  0.0& 65.3 & 0.0 & 65.3& 0.0 \\
 Continual FT &51.2 &	-18.2& 	58.5& 	-6.8 	& 67.2 &	1.9  \\
 \midrule
 LwF\cite{lwf}~ &50.0 &	-19.4& 	49.2& 	-16.1 &	46.5 &	-18.8  \\
 LwF-VR\cite{lwfvr}~ &60.1 &	-9.3 &	62.5 &	-2.8 	& 66.9 &	1.6  \\
 WiSE-FT\cite{wiseft}~&57.7 &	-11.7& 	63.7& 	-1.6 	& 71.8 &	6.5  \\
 ZSCL\cite{zscl} ~ &65.3 &	-4.1 &	64.4 &	-0.9 	& 67.4 &	2.1  \\
 MoE\cite{moe}~ &\underline{68.9} &	\underline{-0.5} &	\underline{71.4} &	\underline{6.1} 	& \underline{76.1} &	\underline{10.8}  \\
 \cellcolor{gray!30}MulKI(ours)  \cellcolor{gray!30}&\textbf{69.5}\cellcolor{gray!30}&	\textbf{0.1}\cellcolor{gray!30}&\textbf{72.4}\cellcolor{gray!30}&\textbf{7.1}\cellcolor{gray!30}	  &\textbf{77.2}\cellcolor{gray!30}	&\textbf{11.9}\cellcolor{gray!30}  \\
\bottomrule
\end{tabular}%
\end{table}
\subsection{Experiments on Multi-domain Task Incremental Learning Setting}
\textit{1) Competitors:} We then evaluate our methods on the setting of multi-domain task incremental learning. Methods (11)-(15) above and (16) WiSE-FT\cite{wiseft} are selected as competitors. In addition, we also add a special competitor ``Full Fine-tune", which means all the tasks are trained jointly. We report ``Transfer", ``Avg." and ``Last" metrics in this setting. \par
\textit{2) Results:}
Table \ref{mtil_order1} reports the detailed results of our proposed MulKI and competitors on MTIL benchmark under Order I. Additional results for Order-II are provided in Table \ref{mtil_order2}. As can be observed in both tables, our method beats all competitors on all three metrics. When Order I is adopted, the ``Transfer" metric of our proposed method is 70.1\%, which is 1.2\% higher than the best competitor MoE and even 0.7\% higher than the original CLIP model. Notably, when Order II is adopted, our proposed method still outperforms the original CLIP model on ``Transfer" metric, while all competitors demonstrate declined performance compared with the original CLIP model on both Order I and II.\par
These results demonstrate the superiority of our proposed method for its capability in effectively mitigating generalization forgetting and catastrophic forgetting while accumulating new knowledge.

\subsection{Few-shot Experiments on Multi-domain Task Incremental Learning Setting}
\textit{1) Settings and Competitors:} 
In this section, we report the results of few-shot experiments on Multi-domain Task Incremental Learning. We adopt a five-shot setting, which limits the model to access only 5 samples per class.\par
We evaluate our methods on few-shot MTIL. Methods (11)-(16) above except for iCaRL\cite{icarl} are selected as competitors. Additionally, we select two special competitors, including ``CLIP Zero-shot" and ``Continual FT", both of them are implemented under few-shot setting. We report ``Transfer", ``Avg." and ``Last" metrics in this setting. \par

\textit{2) Results: }
The comparison results of Order I and II under few-shot setting are shown in Table \ref{few1} and Table \ref{few2}, our proposed method outperforms all competitors on all three metrics. Besides, our method demonstrates a higher improvement than other competitors compared with full-shot setting. As can be observed in \ref{few1}, when Order I is adopted, the ``Transfer" metric of MulKI is 69.5\%, which is 0.6\% higher than the best competitor MoE, and 0.1\% higher than the original CLIP model even with very limited access of training samples. When Order II is adopted, our proposed method still outperforms all competitors at least by 0.5\%, 1.9\% and 1.1\% in terms of three metrics. \par
These results demonstrate that our proposed method is capable of addressing forgetting issues while accumulating new knowledge even with very few samples.
\begin{table}[t]
\centering
\caption{Performance (\%) Comparison of state-of-the-art CL methods on MTIL benchmark in Order II under few-shot setting
}\label{few2}
\begin{tabular}{@{}lcc|cc|cc@{}}
\toprule
Method & Transfer & $\Delta$ & Avg. & $\Delta$ & Last & $\Delta$  \\ \midrule
 CLIP Zero-shot\cite{clip}  &65.4& 	0.0 	&65.3 	&0.0 &65.3 &0.0  \\
 Continual FT &49.2& 	-16.2& 	49.0& 	-16.3&43.8 &-21.5  \\
 \midrule
 LwF\cite{lwf}~ &48.6& 	-16.8& 	54.7& 	-10.6&55.9 &-9.4  \\
 LwF-VR\cite{lwfvr}~&54.4& 	-11.0& 	60.1& 	-5.2 &63.2 &-2.1    \\
 WiSE-FT\cite{wiseft}~ &53.7& 	-11.7& 	55.1& 	-10.2&52.5 &-12.8  \\
 ZSCL\cite{zscl} ~ &62.8& 	-2.6 &	67.6 &	2.3 &72.1 &6.8  \\
 MoE\cite{moe}~ &\underline{64.7}& 	\underline{-0.7} &	\underline{69.5} &	\underline{4.2} &\underline{75.6} &\underline{10.3}  \\
 \cellcolor{gray!30}MulKI(ours)  \cellcolor{gray!30}&\textbf{65.2}\cellcolor{gray!30}& 	\textbf{-0.2}\cellcolor{gray!30}&	\textbf{71.4}\cellcolor{gray!30}&	\textbf{6.1} \cellcolor{gray!30}&\textbf{76.7} \cellcolor{gray!30}&\textbf{11.4}\cellcolor{gray!30}  \\
\bottomrule
\end{tabular}%
\end{table}

\begin{table*}[t]
    \caption{Ablation studies with respect to each component of MulKI on MTIL benchmark in Order I}\label{ablation}
    \scriptsize
    \centering
    \setlength{\tabcolsep}{1.2mm}{
    \begin{tabular}{l|p{10mm}<{\centering}p{9mm}<{\centering}p{9mm}<{\centering}p{9mm}<{\centering}p{9mm}<{\centering}p{9mm}<{\centering}||p{9mm}<{\centering}p{9mm}<{\centering}|p{9mm}<{\centering}p{9mm}<{\centering}|p{9mm}<{\centering}p{10mm}<{\centering}}
        \toprule
        & WC
        & WE
        & CSA 
        & FD 
        & IRD 
        & IDD
        & Transfer
        & impr.
        & Avg.
        & impr.
        & Last
        & impr.
        \\
        \midrule
        Zero-shot& - & - & - & - & - & - &69.4 &	0.0 &	65.3& 	0.0 &	65.3 &	0.0      \\
        Continual-FT& - & - & - & - & - & - &44.6 &-24.8 	&55.9 &	-9.4 &	77.3 &	12.0    \\
        \midrule
        Only FD&   &   &   &   &   &\Checkmark   &63.9 &	-5.5 &	72.5& 	7.2 &	82.9 &	17.6      \\
        Only IRD&   &   &   &   &\Checkmark   &   &66.5 &	-2.9 &	75.2& 	9.9 &	\underline{85.1} &	\underline{19.8}      \\
        Only IDD&   &   &   &\Checkmark   &   &   &64.5 &	-4.9 &	73.8& 	8.5 &	84.2 &	18.9        \\
        \midrule
        Only MDD&  &  &  &\Checkmark   &\Checkmark   &\Checkmark   &67.1 &	-2.3&75.9 &	10.6 &	85.0 &	19.7 \\
        w/o WE\&WC&   &   &\Checkmark   &\Checkmark   &\Checkmark   &\Checkmark   &67.4 &	-2.0 	&76.1 &	10.8 &	\textbf{85.3} &	\textbf{20.0}\\
        w/o WE&   &\Checkmark  &\Checkmark  &\Checkmark  &\Checkmark  &\Checkmark &\underline{69.2} &	\underline{-0.2}&\underline{77.2} &	\underline{11.9} &	\textbf{85.3} &	\textbf{20.0} \\
        \cellcolor{gray!30}MulKI(ours)\cellcolor{gray!30}&\Checkmark  \cellcolor{gray!30}&\Checkmark \cellcolor{gray!30}&\Checkmark\cellcolor{gray!30}&\Checkmark \cellcolor{gray!30}&\Checkmark \cellcolor{gray!30}&\Checkmark\cellcolor{gray!30}&\textbf{70.1} \cellcolor{gray!30}&\textbf{0.7} \cellcolor{gray!30}&\textbf{77.3} \cellcolor{gray!30}&	\textbf{12.0} \cellcolor{gray!30}&\underline{85.1} \cellcolor{gray!30}&\underline{19.8}\cellcolor{gray!30} \\
        \bottomrule
    \end{tabular}
    }
\end{table*}

\begin{table}[t]
\centering
\caption{Performance (\%) Comparison of different weighting strategies on MTIL in Order I}\label{weight}
    \resizebox{170pt}{!}{
    \begin{tabular}{@{}l|ccc@{}}
        \toprule
         Method & Transfer & Avg. & Last    \\ \midrule
         Only $C_0$  &\underline{70.0}&76.9&84.5   \\
         Only $C_{i-1}$ &69.7&76.9&84.7  \\
         Average Weighting~ &69.9&\underline{77.1}&\underline{84.8}   \\
         \cellcolor{gray!30}Our Weighting~\cellcolor{gray!30}&\textbf{70.1}\cellcolor{gray!30}&\textbf{77.3}\cellcolor{gray!30}&\textbf{85.1}\cellcolor{gray!30}     \\
        \bottomrule
    \end{tabular}
    }
\end{table}

\begin{table}[t]
    \caption{Ablation study of Extended Weight Ensemble on CIFAR-100 }\label{ewe}
        \centering
        \resizebox{0.9\linewidth}{!}{
	\begin{tabular}{lcc|cc|cc}
            \toprule 
            \multirow{2}{*}{Method} & \multicolumn{2}{c}{10 step} & \multicolumn{2}{c}{20 step} & \multicolumn{2}{c}{50 step} \\
            \cline{2-7} 
            \noalign{\smallskip}
			 & 	Avg. & Last &Avg. & Last &Avg. &Last  \\
		    \midrule
            WE \cite{zscl}  &86.77 &79.93 &85.19 &76.72 &\textbf{85.00} &75.17 \\
            EWE(ours)       \cellcolor{gray!30}&\textbf{87.17}\cellcolor{gray!30}&\textbf{80.74}\cellcolor{gray!30}&\textbf{85.75}\cellcolor{gray!30}&\textbf{78.50}\cellcolor{gray!30}&84.93\cellcolor{gray!30}&\textbf{75.47}\cellcolor{gray!30}\\
		\bottomrule
	\end{tabular}
        }
\end{table}

\subsection{Ablation Studies}

To evaluate the impact of different components in MulKI, we conduct ablation studies on the MTIL benchmark, as shown in Table \ref{ablation}. We consider the following components: WE and WC apply weight integrations in parameter space, CSA adds cross-modal symmetric alignment loss, and MDD introduces multi-level knowledge distillation loss. Additionally, the three components of MDD are evaluated separately. \par

\textbf{\textit{1) The impact of Model Connection: }}
We conduct ablation experiments to evaluate the roles of WC and WE in the overall model. From Table \ref{ablation}, it is observed that the WE method improves the model's performance for all three metrics. WC method enhances the performance of ``Transfer" and ``Avg." metrics, with an improvement of 0.9\% and 0.3\% respectively. However, the ``Last" metric declines by utilizing WC. The reason this decline occurs is that directly applying WC, which is effectively a regularization loss, to the model weights significantly enhances model’s stability thereby mitigating forgetting even better, but at the expense of model’s plasticity thus damaging the ``Last” metric. \par

\textbf{\textit{2) The impact of Cross-modal Symmetric Alignment: }} We conduct ablation experiments to evaluate the role of the Coss-modal Symmetric Alignment module in the overall model. The CSA module constructs good cross-modal relationships in feature space by aligning visual prototypes and texts, thus brings overall improvement to the model. It is observed in Table \ref{ablation} that the CSA module enhances the performance of all three metrics, with an improvement of 0.3\%, 0.2\% and 0.3\% in three metrics respectively. Although the CSA module does not involve knowledge distillation between teachers and student models, it boosts the accumulation of knowledge in student model for new distributions by ensuring good feature space relationships. \par

\textbf{\textit{3) The impact of Multilevel Knowledge Distillation: }} We conduct ablation experiments on each component of the multi-level dual-teacher knowledge distillation (MDD) module to evaluate their individual contributions. Table \ref{ablation} shows the results for FD, IRD, and IDD modules, as well as the overall MDD module. It is observed that the performance achieved by the three modules individually exhibit significant decrease compared to the whole MDD module. The FD module performs the worst since it lacks inter-modality interaction and fine-grained information distillation. The experiments on IRD and IDD modules each show that IRD module achieves better performance, which even shows consistent performance with the complete MDD module on the ``Last" metric, but IRD module exhibits too much plasticity with a lack on stability. All three distillation modules combine a complete MDD module, achieving a balance between stability and plasticity at the cost of slightly sacrificing plasticity. \par


\textbf{\textit{4) The impact of Similarity-based Weighting Strategy: }}we compare the proposed similarity-based weighting strategy with various weighting strategies on the MTIL benchmark to evaluate its effectiveness, including distillation using only the $C_0$ model, only the $C_{i-1}$ model, and average weighting with both models. Table \ref{weight} presents these comparison results.\par
\begin{figure}
	\begin{center}
		\includegraphics[scale=0.85]{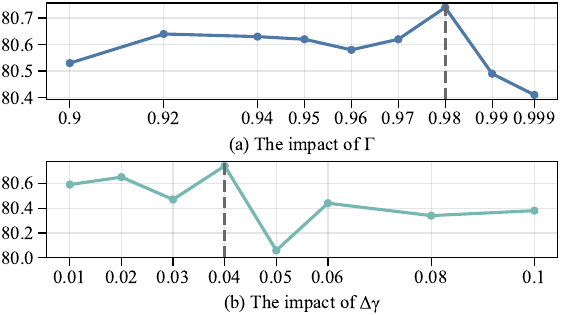}
	\end{center}
	\caption{The impact of the upperbound of $\gamma$ and its increasing step for CIFAR-100 on 10-step setting. We report the ``Last" metric. The chosen hyper-parameters are marked with dash lines. (a) The impact of $\Gamma$ when $\Delta \gamma$ is set as 0.04. (b) The impact of $\Delta \gamma$ when $\Gamma$ is set as 0.98.}
	\label{hyper}
\end{figure}
\begin{figure*}[htbp]
      \centering
      \subfigure[CLIP]{
           \fbox{\includegraphics[scale=0.15]{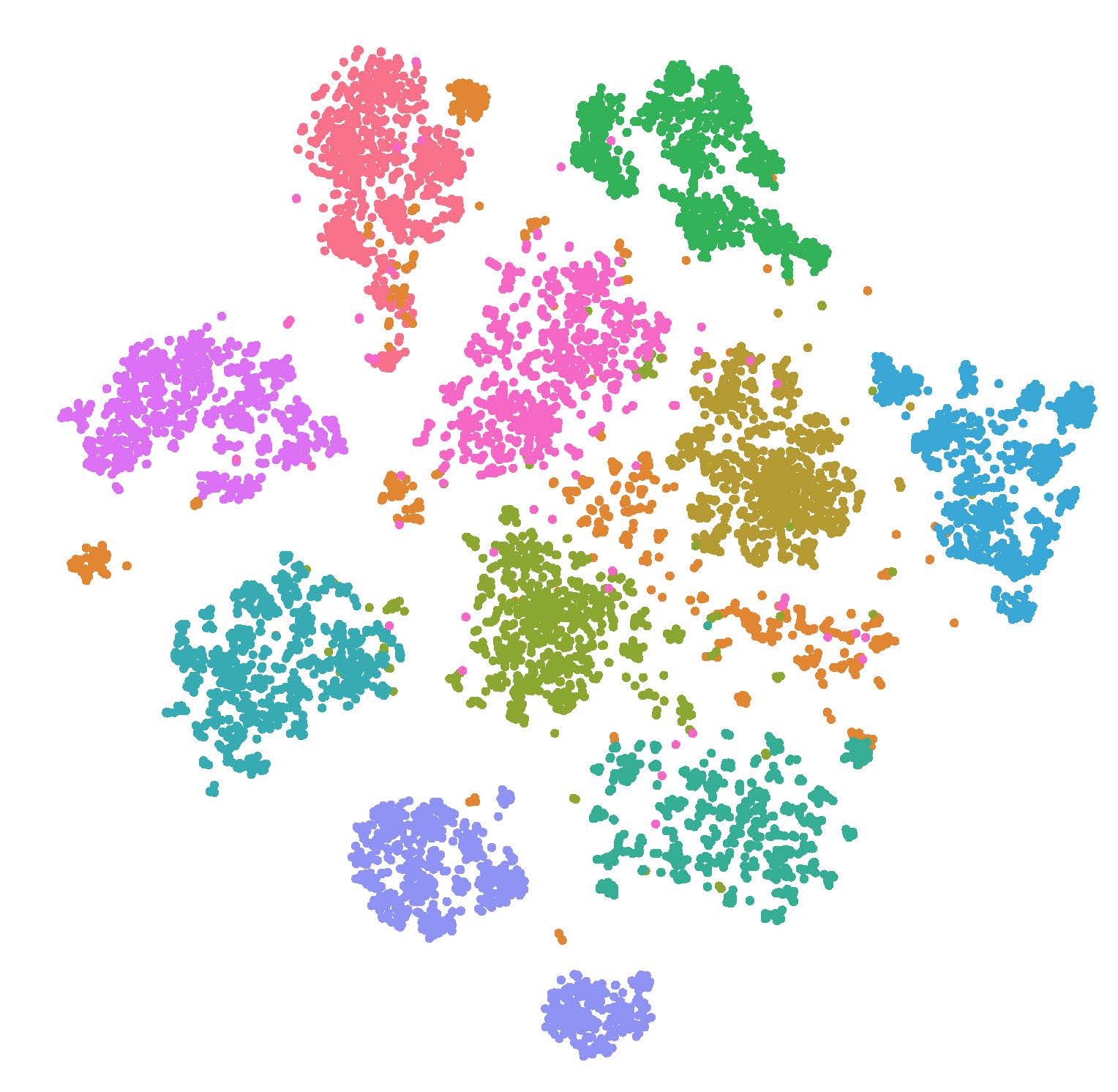}\label{vis1}}
      }\hspace*{-5pt}
      \subfigure[ZSCL]{
           \fbox{\includegraphics[scale=0.15]{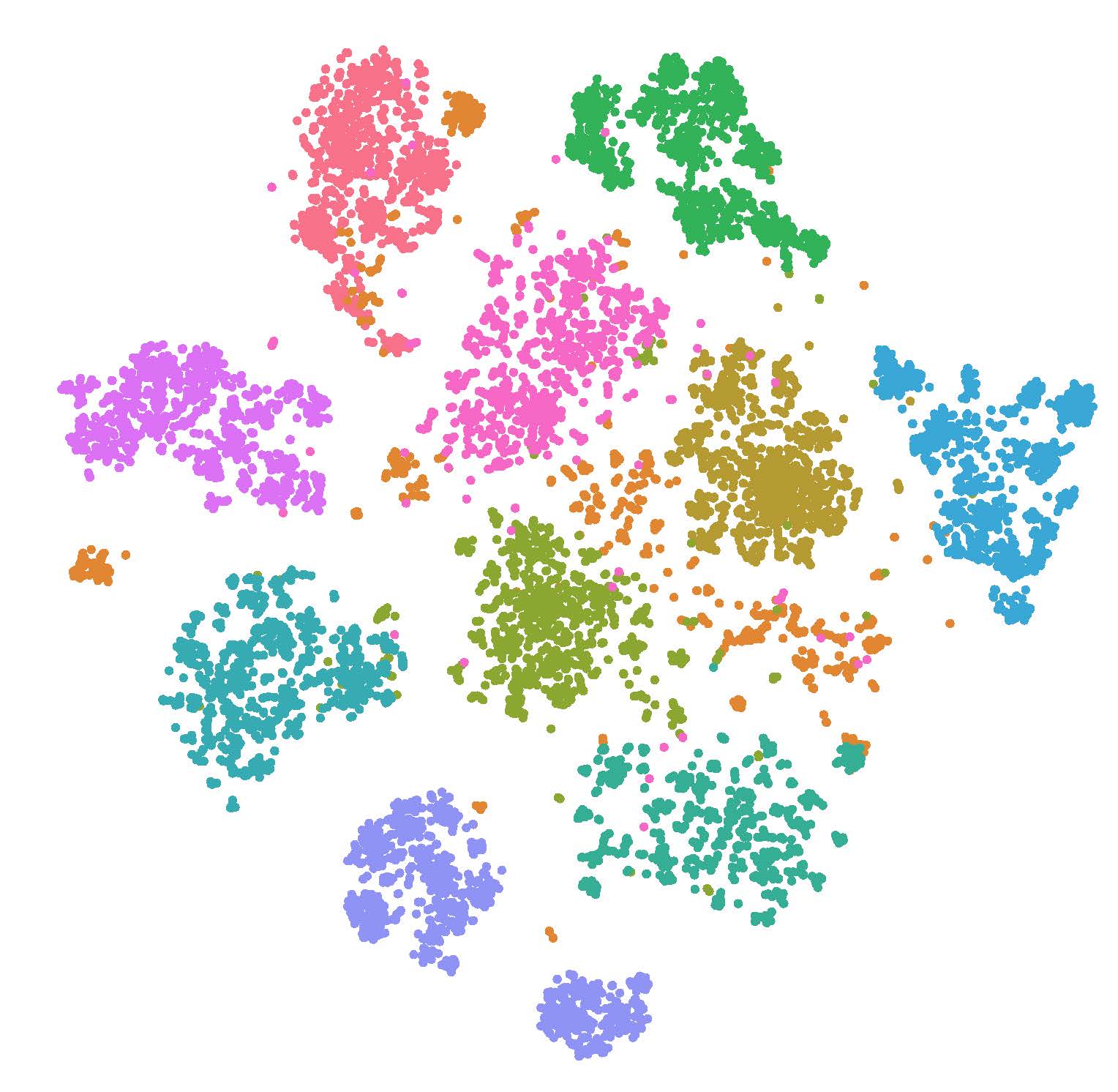}\label{vis2}}
      }\hspace*{-5pt}
      \subfigure[MoE]{
           \fbox{\includegraphics[scale=0.15]{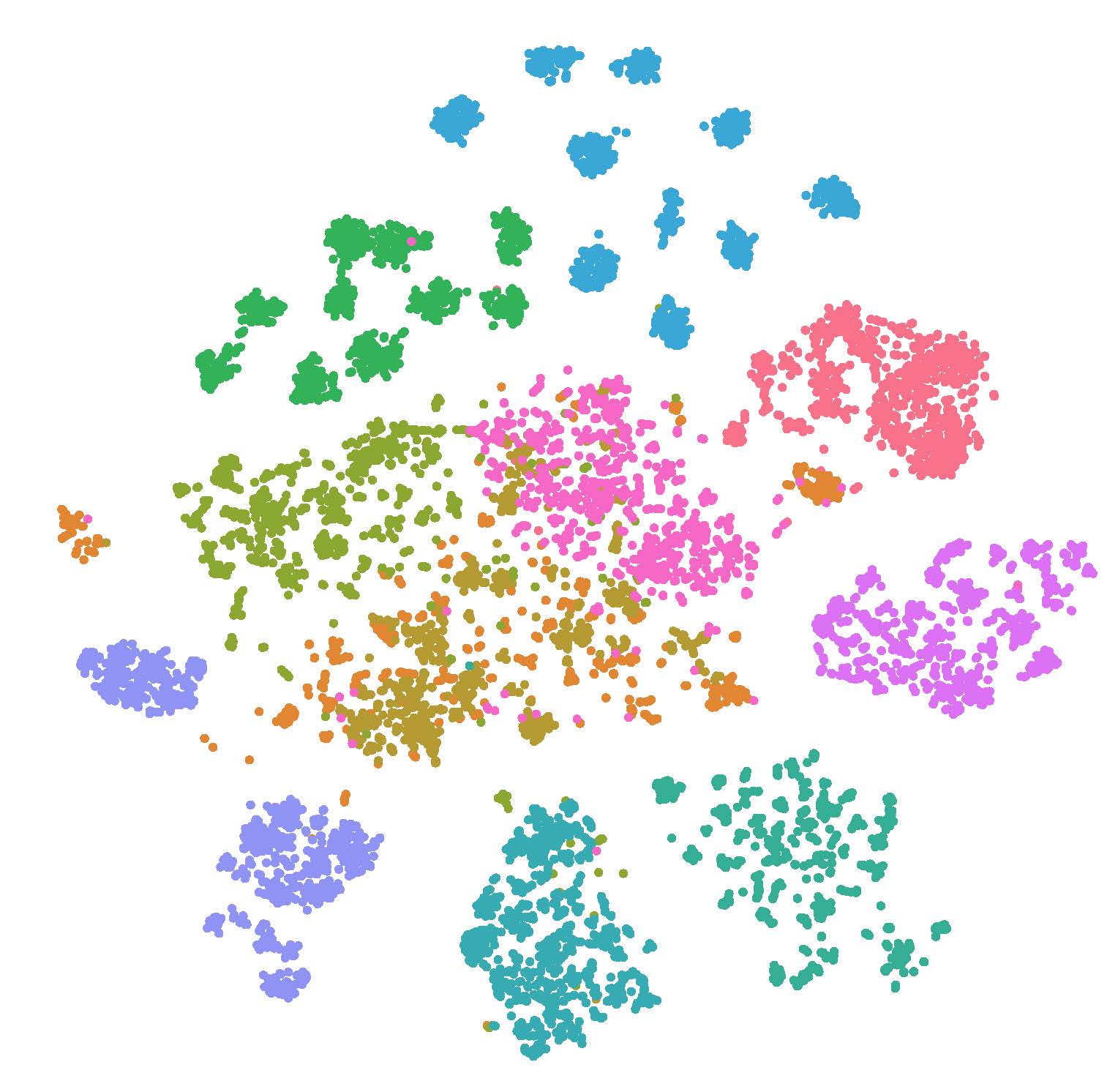}\label{vis3}}
      }\hspace*{-5pt}
      \subfigure[Few-shot MulKI]{
           \fbox{\includegraphics[scale=0.15]{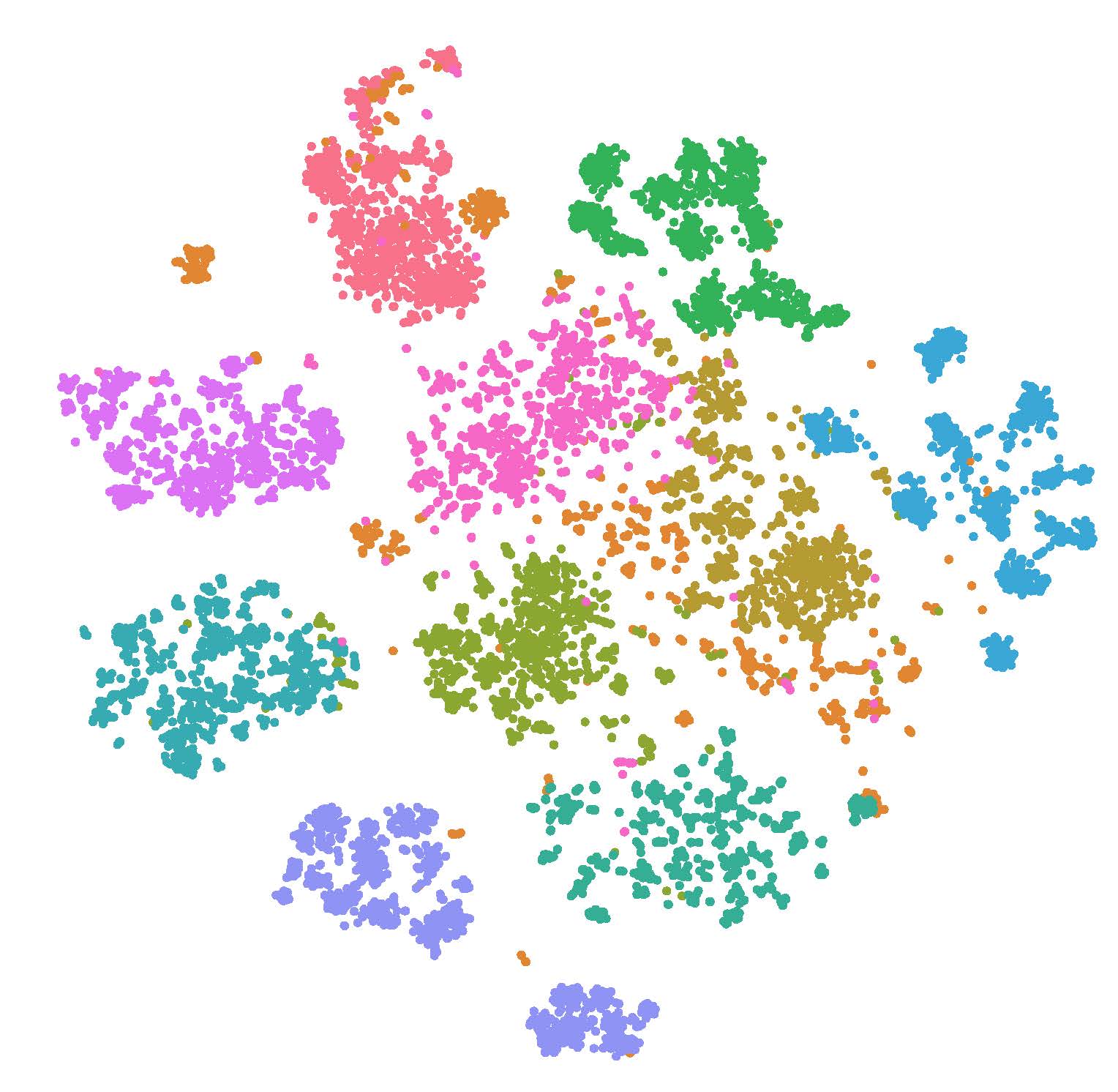}\label{vis4}}
      }\hspace*{-5pt}
      \subfigure[Full-shot MulKI]{
           \fbox{\includegraphics[scale=0.15]{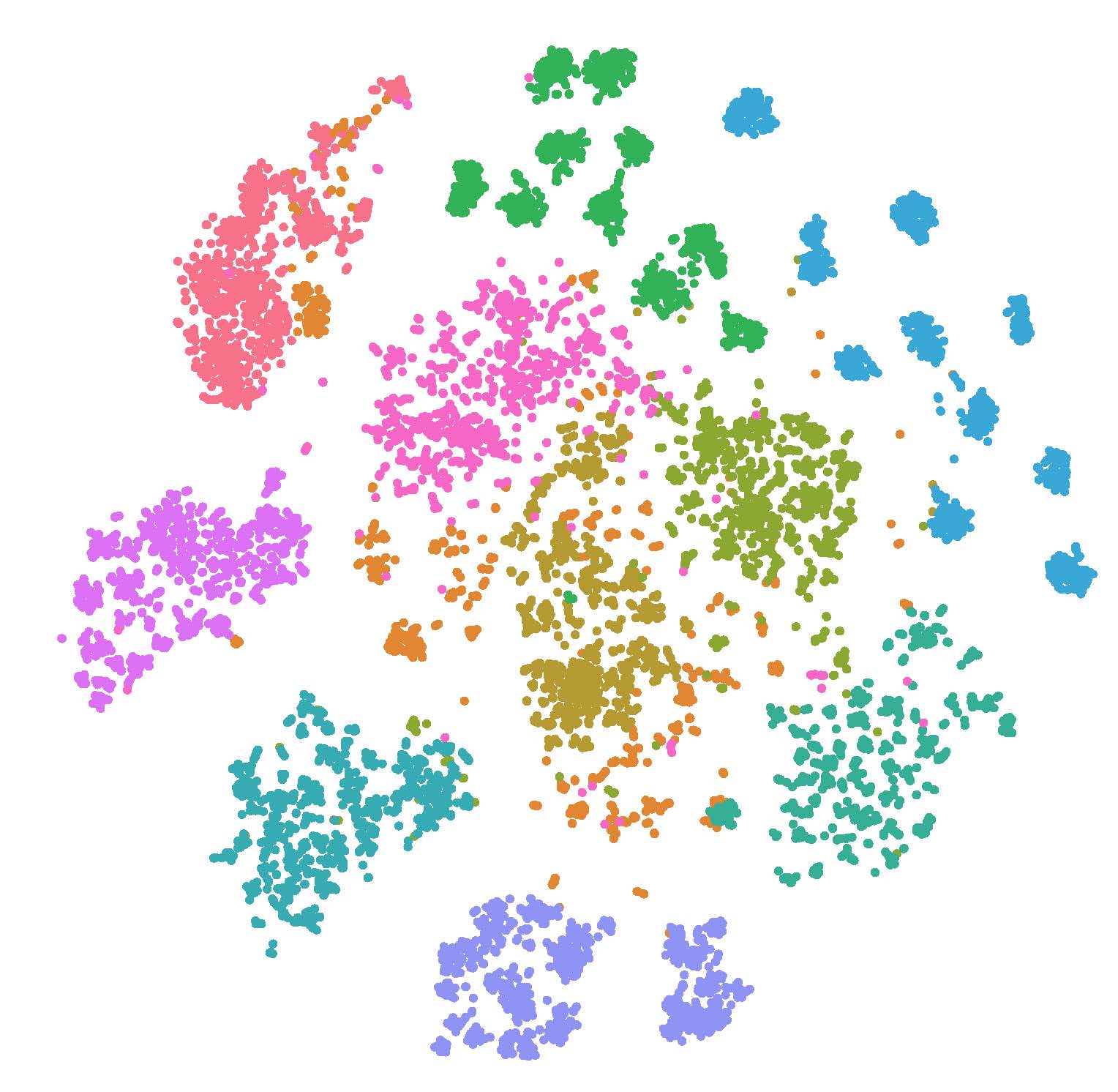}\label{vis5}}
      } \hspace*{-10pt}
      \subfigure{
           \fbox{\includegraphics[scale=0.496]{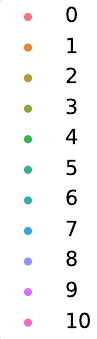}}
      }  
      \caption{t-SNE visualization of features of (a) CLIP, (b) ZSCL, (c) MoE, (d) Few-shot MulKI and (e) Full-Shot MulKI for MTIL benchmark under Order I. ZSCL and our MulKI under few-shot setting maintain the similar feature distribution to the original CLIP, while MoE significantly distorts the feature space. 0-10 denote the task ids in Order I.}\label{vis}
\end{figure*}
\begin{figure*}[htbp]
      \centering
      \subfigure[Original]{
           \includegraphics[scale=0.35]{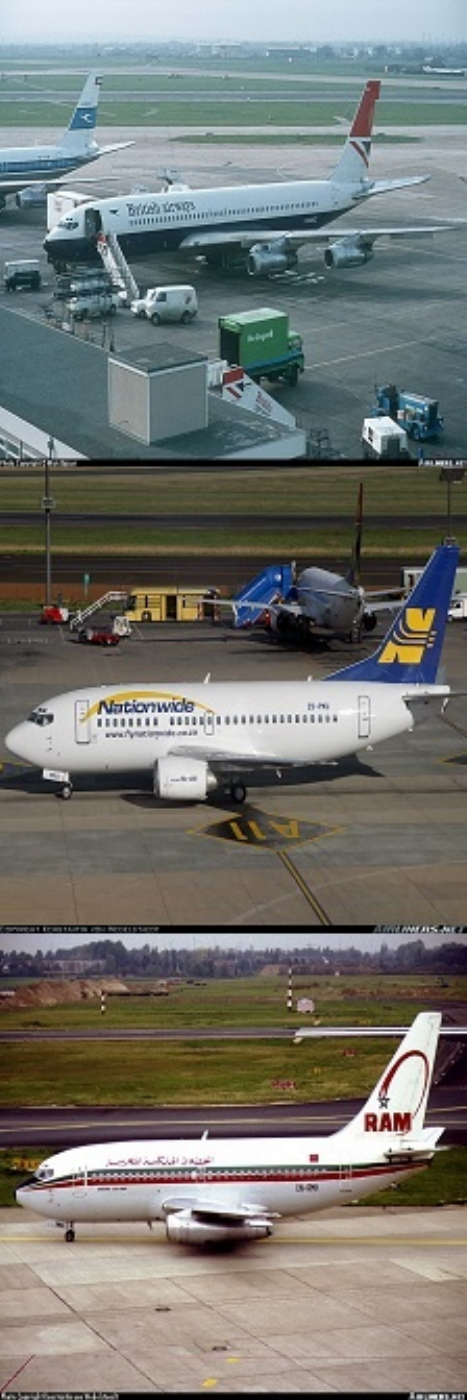}\label{cam1}
      }\hspace*{-11pt}
      \subfigure[CLIP]{
           \includegraphics[scale=0.35]{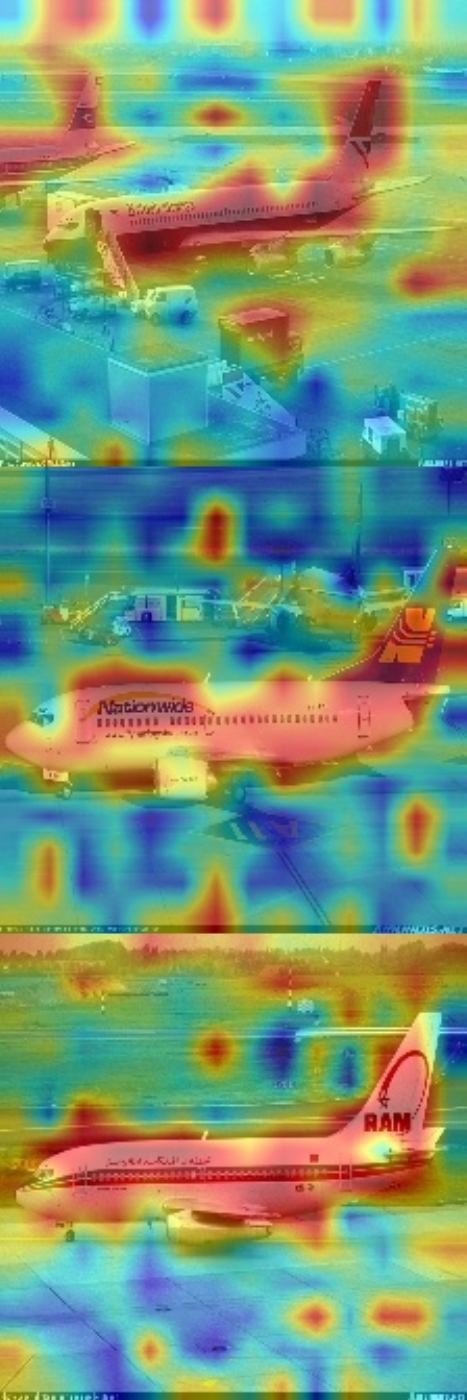}\label{cam2}
      }\hspace*{-11pt}
      \subfigure[ZSCL]{
           \includegraphics[scale=0.35]{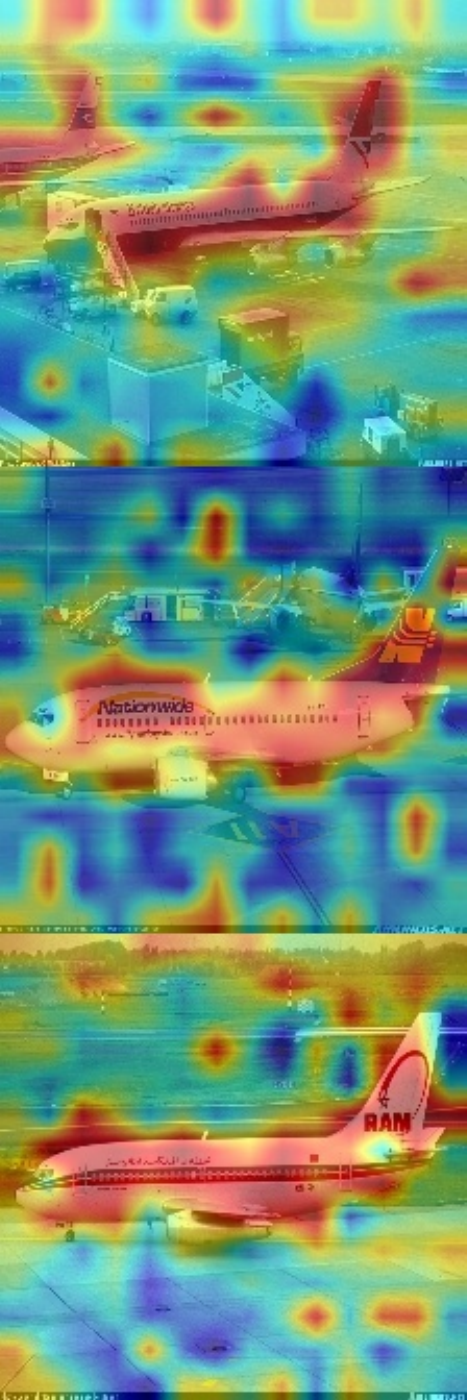}\label{cam3}
      }\hspace*{-11pt}
      \subfigure[MoE]{
           \includegraphics[scale=0.35]{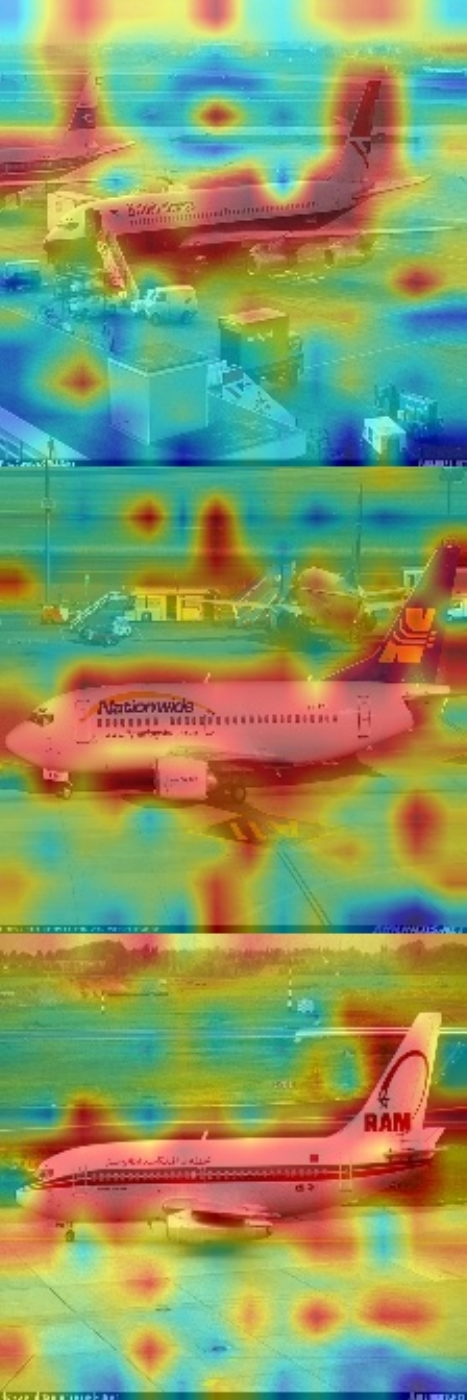}\label{cam4}
      }\hspace*{-11pt}
      \subfigure[Few-shot MulKI]{
           \includegraphics[scale=0.35]{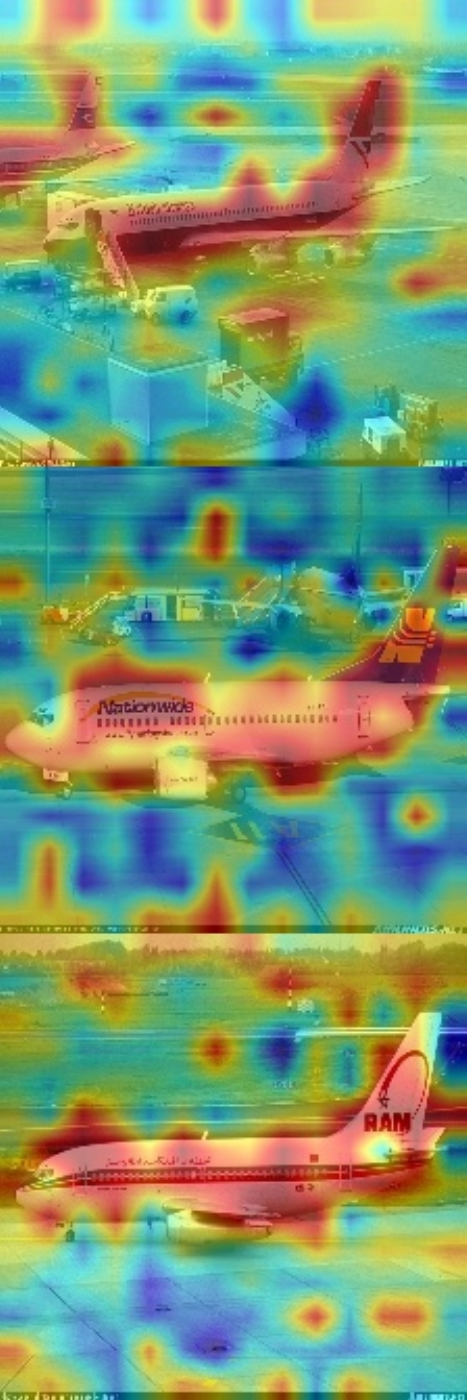}\label{cam5}
      }\hspace*{-11pt}
      \subfigure[Full-shot MulKI]{
           \includegraphics[scale=0.35]{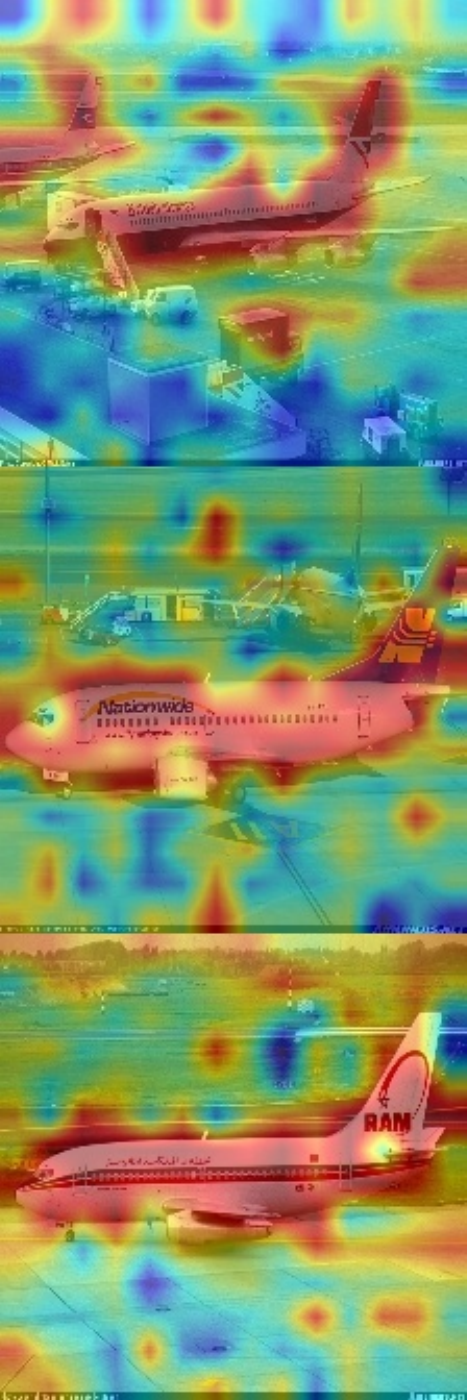}\label{cam6}
      }

      \caption{Grad-CAM visualization results of (b) CLIP\cite{clip}, (c) ZSCL\cite{zscl}, (d) MoE\cite{moe}, (e) Few-shot MulKI and (f) Full-Shot MulKI for test samples of Aircraft with the last model. Our MulKI is competitive with MoE, which is particularly designed for enhancing long-term memory, even under few-shot setting.}\label{cam}
\end{figure*}
Distilling using only the $C_0$ model retains the zero-shot capability of the CLIP pre-trained model but compromises downstream task knowledge, resulting in poor ``Last" metric performance. Conversely, using only the $C_{i-1}$ model preserves downstream task knowledge but sacrifices zero-shot capabilities, leading to poor ``Transfer" metric performance. Average weighting, which combines both models, performs better than ``Only $C_{i-1}$" on all metrics but worse than ``Only $C_0$" on the ``Transfer" metric. Although it alleviates both forgetting issues, average weighting doesn't account for individual samples, resulting in suboptimal outcomes. In contrast, our weighting strategy assigns distillation weights to individual samples, balancing stability and plasticity, and achieving competitive performance across all metrics.

\textbf{\textit{5) The impact of Extended Weight Ensemble Strategy: }}We conduct ablation experiments to evaluate the roles of EWE strategy in the overall model for CIFAR-100. From Table \ref{ewe}, it is observed that model with EWE strategy performs better than that with WE in all three cases, except for the ``Avg." metric in ``50-step" case. Particularly, in case of ``20-step", EWE reaches an improvement of 0.56\% and 1.78\% in two metrics. However, EWE fails on TinyImageNet and MTIL. It is because EWE works by enhancing the model’s stability, EWE helps CIFAR-100 to reach a better stability-plasticity balance that WE is not capable of, while WE already enables a reasonable stability-plasticity balance for TinyImageNet and MTIL, EWE offers too much stability.

\textbf{\textit{6) The impact of $\gamma$ and its step: }}The hyper-parameter $\gamma$ and its step size are crucial for constructing prototypes in formula (\ref{pro_cons}). We perform ablation experiments on CIFAR-100 with a 10-step setting to assess the impact of the upper bound of $\gamma$ (denoted as $\Gamma$) and its step size ($\Delta \gamma$). Figure \ref{hyper} shows the effect of $\Gamma$ and $\Delta \gamma$ on the ``Last" metric. Performance declines with other hyper-parameter choices, indicating that only appropriate $\Gamma$ and $\Delta \gamma$ can achieve reasonable construction of prototypes. A too large or too small $\Delta \gamma$ hinders prototype construction, while an improper $\Gamma$ fails to preserve the well-constructed prototypes. The chosen hyper-parameters in Fig. \ref{hyper} effectively facilitate and preserve well-constructed prototypes.

\subsection{Visualization Analysis}
\textbf{\textit{1) t-SNE Visualization for Features: }}
To further verify the performance of our MulKI network, we conduct visualization analysis to visualize the features of the test set of MTIL for original CLIP, ZSCL, MoE and our MulKI, in which MulKI includes few-shot and full-shot settings, after the whole MTIL is trained. Figure \ref{vis} provides the t-SNE visualization results. It is observed that ZSCL and our MulKI under few-shot setting maintain the similar feature distribution to the original CLIP, and our MulKI under full-shot setting to distorts the feature space at an acceptable level, while MoE significantly distorts the feature space. This further demonstrates the effectiveness of our MulKI, as although it is not particularly designed for maintaining the feature space like ZSCL, which utilize semantically rich reference dataset to preserve the feature space, our MulKI achieves a reasonable preservation of the feature space, especially under few-shot setting. \par
\textbf{\textit{2) Grad-CAM Visualization for Features: }}
 To verify the overall performance and the interpretability of our MulKI network, we visualize the attention maps from the last ViT transformer block. We use the final model of MTIL to visualize the test samples from the first dataset Aircraft, which verifies the model's long-term memory ability additionally. We choose three test samples from distinct classes. Figure \ref{cam} provides the Grad-CAM visualization results. It is observed that our proposed MulKI achieves competitive visulization result with MoE, which is particularly designed for enhancing model's long-term memory, even under the few-shot setting.\par

\section{Conclusion}

In this paper, We have proposed a Multi-Stage Knowledge Integration method to enhance continual learning in vision-language models by emulating the human learning process. Our approach includes four stages: Eliciting Ideas, Adding New Ideas, Distinguishing Ideas, and Making Connections. In these stages, knowledge from different modalities is elicited and used to build fine-grained intra- and inter-modality relationships with prototypes. We introduce new data and information from two teacher models at three levels, then adaptively weighting the knowledge. We then integrate preceding and new knowledge from intra- and inter-task perspectives by establishing connections based on model weights. Our method fully leverages multimodal information from the training dataset without additional data, creating an outstanding student model by comprehensively using two teacher models. Extensive experiments on traditional class-incremental learning and novel multi-domain task-incremental learning settings demonstrate the effectiveness of our method.

\ifCLASSOPTIONcaptionsoff
  \newpage
\fi



%

\bibliographystyle{IEEEtran}
\bibliography{mulki}

\begin{thebibliography}{10}
\providecommand{\url}[1]{#1}
\csname url@samestyle\endcsname
\providecommand{\newblock}{\relax}
\providecommand{\bibinfo}[2]{#2}
\providecommand{\BIBentrySTDinterwordspacing}{\spaceskip=0pt\relax}
\providecommand{\BIBentryALTinterwordstretchfactor}{4}
\providecommand{\BIBentryALTinterwordspacing}{\spaceskip=\fontdimen2\font plus
\BIBentryALTinterwordstretchfactor\fontdimen3\font minus \fontdimen4\font\relax}
\providecommand{\BIBforeignlanguage}[2]{{%
\expandafter\ifx\csname l@#1\endcsname\relax
\typeout{** WARNING: IEEEtran.bst: No hyphenation pattern has been}%
\typeout{** loaded for the language `#1'. Using the pattern for}%
\typeout{** the default language instead.}%
\else
\language=\csname l@#1\endcsname
\fi
#2}}
\providecommand{\BIBdecl}{\relax}
\BIBdecl

\bibitem{zscl}
Z.~Zheng, M.~Ma, K.~Wang, Z.~Qin, X.~Yue, and Y.~You, ``Preventing zero-shot transfer degradation in continual learning of vision-language models,'' in \emph{Proc. IEEE/CVF Conf. Comput. Vis. Pattern Recognit. (CVPR)}, 2023, pp. 19\,125--19\,136.

\bibitem{jia2021scaling}
C.~Jia, Y.~Yang, Y.~Xia, Y.-T. Chen, Z.~Parekh, H.~Pham, Q.~Le, Y.-H. Sung, Z.~Li, and T.~Duerig, ``Scaling up visual and vision-language representation learning with noisy text supervision,'' in \emph{Proc. Int. Conf. Mach. Learn.}\hskip 1em plus 0.5em minus 0.4em\relax PMLR, 2021, pp. 4904--4916.

\bibitem{li2022blip}
J.~Li, D.~Li, C.~Xiong, and S.~Hoi, ``Blip: Bootstrapping language-image pre-training for unified vision-language understanding and generation,'' in \emph{Proc. Int. Conf. Mach. Learn.}\hskip 1em plus 0.5em minus 0.4em\relax PMLR, 2022, pp. 12\,888--12\,900.

\bibitem{yao2021filip}
L.~Yao, R.~Huang, L.~Hou, G.~Lu, M.~Niu, H.~Xu, X.~Liang, Z.~Li, X.~Jiang, and C.~Xu, ``Filip: Fine-grained interactive language-image pre-training,'' \emph{arXiv preprint arXiv:2111.07783}, 2021.

\bibitem{zy}
Y.~Zhang, Z.~Ji, D.~Wang, Y.~Pang, and X.~Li, ``User: Unified semantic enhancement with momentum contrast for image-text retrieval,'' \emph{IEEE Trans. on Image Process.}, 2024, doi:{\color{blue} \href{https://doi.org/10.1109/TIP.2023.3348297}{10.1109/TIP.2023.3348297}}.

\bibitem{catas}
M.~McCloskey and N.~J. Cohen, ``Catastrophic interference in connectionist networks: The sequential learning problem,'' in \emph{Psychology of learning and motivation}.\hskip 1em plus 0.5em minus 0.4em\relax Elsevier, 1989, vol.~24, pp. 109--165.

\bibitem{lwfvr}
Y.~Ding, L.~Liu, C.~Tian, J.~Yang, and H.~Ding, ``Don't stop learning: Towards continual learning for the clip model,'' \emph{arXiv preprint arXiv:2207.09248}, 2022.

\bibitem{moe}
J.~Yu, Y.~Zhuge, L.~Zhang, P.~Hu, D.~Wang, H.~Lu, and Y.~He, ``Boosting continual learning of vision-language models via mixture-of-experts adapters,'' in \emph{Proc. IEEE/CVF Conf. Comput. Vis. Pattern Recognit. (CVPR)}, 2024, pp. 23\,219--23\,230.

\bibitem{sprompt}
Y.~Wang, Z.~Huang, and X.~Hong, ``S-prompts learning with pre-trained transformers: An occam’s razor for domain incremental learning,'' in \emph{Proc. Int. Conf. Neural Inf. Process. Syst. (NeurIPS)}, vol.~35, 2022, pp. 5682--5695.

\bibitem{imagenet}
O.~Russakovsky, J.~Deng, H.~Su, J.~Krause, S.~Satheesh, S.~Ma, Z.~Huang, A.~Karpathy, A.~Khosla, M.~Bernstein \emph{et~al.}, ``Imagenet large scale visual recognition challenge,'' \emph{Int. J. Comput. Vis.}, vol. 115, pp. 211--252, 2015.

\bibitem{platonic}
M.~Huh, B.~Cheung, T.~Wang, and P.~Isola, ``The platonic representation hypothesis,'' \emph{arXiv preprint arXiv:2405.07987}, 2024.

\bibitem{bell1995knowledge}
P.~Bell, E.~A. Davis, and M.~C. Linn, ``The knowledge integration environment: Theory and design,'' 1995.

\bibitem{lwf}
Z.~Li and D.~Hoiem, ``Learning without forgetting,'' \emph{IEEE Trans. Pattern Anal. Mach. Intell.}, vol.~40, no.~12, pp. 2935--2947, 2017.

\bibitem{gdumb}
A.~Prabhu, P.~H. Torr, and P.~K. Dokania, ``Gdumb: A simple approach that questions our progress in continual learning,'' in \emph{Proc. Eur. Conf. Comput. Vis. (ECCV)}.\hskip 1em plus 0.5em minus 0.4em\relax Springer, 2020, pp. 524--540.

\bibitem{latent_replay}
O.~Ostapenko, T.~Lesort, P.~Rodr{\'\i}guez, M.~R. Arefin, A.~Douillard, I.~Rish, and L.~Charlin, ``Continual learning with foundation models: An empirical study of latent replay,'' in \emph{Proc. of the 1st Conf. on Lifelong Learning Agents (CoLLAs)}.\hskip 1em plus 0.5em minus 0.4em\relax PMLR, 2022, pp. 60--91.

\bibitem{icarl}
S.-A. Rebuffi, A.~Kolesnikov, G.~Sperl, and C.~H. Lampert, ``icarl: Incremental classifier and representation learning,'' in \emph{Proc. IEEE/CVF Conf. Comput. Vis. Pattern Recognit. (CVPR)}, 2017, pp. 2001--2010.

\bibitem{der}
P.~Buzzega, M.~Boschini, A.~Porrello, D.~Abati, and S.~Calderara, ``Dark experience for general continual learning: a strong, simple baseline,'' in \emph{Proc. Int. Conf. Neural Inf. Process. Syst. (NeurIPS)}, vol.~33, 2020, pp. 15\,920--15\,930.

\bibitem{dgr}
H.~Shin, J.~K. Lee, J.~Kim, and J.~Kim, ``Continual learning with deep generative replay,'' in \emph{Proc. Int. Conf. Neural Inf. Process. Syst. (NeurIPS)}, vol.~30, 2017.

\bibitem{lgm}
J.~Ramapuram, M.~Gregorova, and A.~Kalousis, ``Lifelong generative modeling,'' \emph{Neurocomputing}, vol. 404, pp. 381--400, 2020.

\bibitem{2018pseudo}
C.~Atkinson, B.~McCane, L.~Szymanski, and A.~Robins, ``Pseudo-recursal: Solving the catastrophic forgetting problem in deep neural networks,'' \emph{arXiv preprint arXiv:1802.03875}, 2018.

\bibitem{gan}
I.~Goodfellow, J.~Pouget-Abadie, M.~Mirza, B.~Xu, D.~Warde-Farley, S.~Ozair, A.~Courville, and Y.~Bengio, ``Generative adversarial nets,'' in \emph{Proc. Int. Conf. Neural Inf. Process. Syst. (NeurIPS)}, vol.~27, 2014.

\bibitem{vae}
D.~P. Kingma and M.~Welling, ``Auto-encoding variational bayes,'' \emph{arXiv preprint arXiv:1312.6114}, 2013.

\bibitem{Memoryawaresynapses}
R.~Aljundi, F.~Babiloni, M.~Elhoseiny, M.~Rohrbach, and T.~Tuytelaars, ``Memory aware synapses: Learning what (not) to forget,'' in \emph{Proc. Eur. Conf. Comput. Vis. (ECCV)}, 2018, pp. 139--154.

\bibitem{ewc}
J.~Kirkpatrick, R.~Pascanu, N.~Rabinowitz, J.~Veness, G.~Desjardins, A.~A. Rusu, K.~Milan, J.~Quan, T.~Ramalho, A.~Grabska-Barwinska \emph{et~al.}, ``Overcoming catastrophic forgetting in neural networks,'' \emph{Proc. Nat. Acad. Sci. USA}, vol. 114, no.~13, pp. 3521--3526, 2017.

\bibitem{lee2017overcoming}
S.-W. Lee, J.-H. Kim, J.~Jun, J.-W. Ha, and B.-T. Zhang, ``Overcoming catastrophic forgetting by incremental moment matching,'' in \emph{Proc. Int. Conf. Neural Inf. Process. Syst. (NeurIPS)}, vol.~30, 2017.

\bibitem{dhar2019learning}
P.~Dhar, R.~V. Singh, K.-C. Peng, Z.~Wu, and R.~Chellappa, ``Learning without memorizing,'' in \emph{Proc. IEEE/CVF Conf. Comput. Vis. Pattern Recognit. (CVPR)}, 2019, pp. 5138--5146.

\bibitem{podnet}
A.~Douillard, M.~Cord, C.~Ollion, T.~Robert, and E.~Valle, ``Podnet: Pooled outputs distillation for small-tasks incremental learning,'' in \emph{Proc. Eur. Conf. Comput. Vis. (ECCV)}.\hskip 1em plus 0.5em minus 0.4em\relax Springer, 2020, pp. 86--102.

\bibitem{lucir}
S.~Hou, X.~Pan, C.~C. Loy, Z.~Wang, and D.~Lin, ``Learning a unified classifier incrementally via rebalancing,'' in \emph{Proc. IEEE/CVF Conf. Comput. Vis. Pattern Recognit. (CVPR)}, 2019, pp. 831--839.

\bibitem{ckdf}
K.~Li, J.~Wan, and S.~Yu, ``Ckdf: Cascaded knowledge distillation framework for robust incremental learning,'' \emph{IEEE Trans. on Image Process.}, vol.~31, pp. 3825--3837, 2022.

\bibitem{lijin}
Z.~Ji, J.~Li, Q.~Wang, and Z.~Zhang, ``Complementary calibration: Boosting general continual learning with collaborative distillation and self-supervision,'' \emph{IEEE Trans. on Image Process.}, vol.~32, pp. 657--667, 2022.

\bibitem{lu2024pamk}
J.~Lu and S.~Sun, ``Pamk: Prototype augmented multi-teacher knowledge transfer network for continual zero-shot learning,'' \emph{IEEE Trans. on Image Process.}, 2024, doi:{\color{blue} \href{https://doi.org/10.1109/TIP.2024.3403053}{10.1109/TIP.2024.3403053}}.

\bibitem{hu2023dense}
Z.~Hu, Y.~Li, J.~Lyu, D.~Gao, and N.~Vasconcelos, ``Dense network expansion for class incremental learning,'' in \emph{Proc. IEEE/CVF Conf. Comput. Vis. Pattern Recognit. (CVPR)}, 2023, pp. 11\,858--11\,867.

\bibitem{ye2023self}
F.~Ye and A.~G. Bors, ``Self-evolved dynamic expansion model for task-free continual learning,'' in \emph{Proc. IEEE/CVF Conf. Comput. Vis. Pattern Recognit. (CVPR)}, 2023, pp. 22\,102--22\,112.

\bibitem{der2021}
S.~Yan, J.~Xie, and X.~He, ``Der: Dynamically expandable representation for class incremental learning,'' in \emph{Proc. IEEE/CVF Conf. Comput. Vis. Pattern Recognit. (CVPR)}, 2021, pp. 3014--3023.

\bibitem{dytox}
A.~Douillard, A.~Ram{\'e}, G.~Couairon, and M.~Cord, ``Dytox: Transformers for continual learning with dynamic token expansion,'' in \emph{Proc. IEEE/CVF Conf. Comput. Vis. Pattern Recognit. (CVPR)}, 2022, pp. 9285--9295.

\bibitem{yoon2017lifelong}
J.~Yoon, E.~Yang, J.~Lee, and S.~J. Hwang, ``Lifelong learning with dynamically expandable networks,'' \emph{arXiv preprint arXiv:1708.01547}, 2017.

\bibitem{expert}
R.~Aljundi, P.~Chakravarty, and T.~Tuytelaars, ``Expert gate: Lifelong learning with a network of experts,'' in \emph{Proc. IEEE/CVF Conf. Comput. Vis. Pattern Recognit. (CVPR)}, 2017, pp. 3366--3375.

\bibitem{vit}
A.~Dosovitskiy, L.~Beyer, A.~Kolesnikov, D.~Weissenborn, X.~Zhai, T.~Unterthiner, M.~Dehghani, M.~Minderer, G.~Heigold, S.~Gelly \emph{et~al.}, ``An image is worth 16x16 words: Transformers for image recognition at scale,'' \emph{arXiv preprint arXiv:2010.11929}, 2020.

\bibitem{l2p}
Z.~Wang, Z.~Zhang, C.-Y. Lee, H.~Zhang, R.~Sun, X.~Ren, G.~Su, V.~Perot, J.~Dy, and T.~Pfister, ``Learning to prompt for continual learning,'' in \emph{Proc. IEEE/CVF Conf. Comput. Vis. Pattern Recognit. (CVPR)}, 2022, pp. 139--149.

\bibitem{dualprompt}
Z.~Wang, Z.~Zhang, S.~Ebrahimi, R.~Sun, H.~Zhang, C.-Y. Lee, X.~Ren, G.~Su, V.~Perot, J.~Dy \emph{et~al.}, ``Dualprompt: Complementary prompting for rehearsal-free continual learning,'' in \emph{Proc. Eur. Conf. Comput. Vis. (ECCV)}.\hskip 1em plus 0.5em minus 0.4em\relax Springer, 2022, pp. 631--648.

\bibitem{coda}
J.~S. Smith, L.~Karlinsky, V.~Gutta, P.~Cascante-Bonilla, D.~Kim, A.~Arbelle, R.~Panda, R.~Feris, and Z.~Kira, ``Coda-prompt: Continual decomposed attention-based prompting for rehearsal-free continual learning,'' in \emph{Proc. IEEE/CVF Conf. Comput. Vis. Pattern Recognit. (CVPR)}, 2023, pp. 11\,909--11\,919.

\bibitem{lora}
E.~J. Hu, Y.~Shen, P.~Wallis, Z.~Allen-Zhu, Y.~Li, S.~Wang, L.~Wang, and W.~Chen, ``Lora: Low-rank adaptation of large language models,'' \emph{arXiv preprint arXiv:2106.09685}, 2021.

\bibitem{zhao2024learning}
C.~Zhao, Y.~Wang, X.~Jiang, Y.~Shen, K.~Song, D.~Li, and D.~Miao, ``Learning domain invariant prompt for vision-language models,'' \emph{IEEE Trans. on Image Process.}, 2024, doi:{\color{blue} \href{https://doi.org/10.1109/TIP.2024.3362062}{10.1109/TIP.2024.3362062}}.

\bibitem{proof}
D.-W. Zhou, Y.~Zhang, J.~Ning, H.-J. Ye, D.-C. Zhan, and Z.~Liu, ``Learning without forgetting for vision-language models,'' \emph{arXiv preprint arXiv:2305.19270}, 2023.

\bibitem{coleclip}
Y.~Li, G.~Pang, W.~Suo, C.~Jing, Y.~Xi, L.~Liu, H.~Chen, G.~Liang, and P.~Wang, ``Coleclip: Open-domain continual learning via joint task prompt and vocabulary learning,'' \emph{arXiv preprint arXiv:2403.10245}, 2024.

\bibitem{clip}
A.~Radford, J.~W. Kim, C.~Hallacy, A.~Ramesh, G.~Goh, S.~Agarwal, G.~Sastry, A.~Askell, P.~Mishkin, J.~Clark \emph{et~al.}, ``Learning transferable visual models from natural language supervision,'' in \emph{Proc. Int. Conf. Mach. Learn.}\hskip 1em plus 0.5em minus 0.4em\relax PMLR, 2021, pp. 8748--8763.

\bibitem{cifar}
A.~Krizhevsky, G.~Hinton \emph{et~al.}, ``Learning multiple layers of features from tiny images,'' \emph{Univ. Toronto, Toronto, ON, Canada, Tech. Rep.}, 2009.

\bibitem{bic}
Y.~Wu, Y.~Chen, L.~Wang, Y.~Ye, Z.~Liu, Y.~Guo, and Y.~Fu, ``Large scale incremental learning,'' in \emph{Proc. IEEE/CVF Conf. Comput. Vis. Pattern Recognit. (CVPR)}, 2019, pp. 374--382.

\bibitem{rspnet}
J.~Rajasegaran, M.~Hayat, S.~Khan, F.~S. Khan, L.~Shao, and M.-H. Yang, ``An adaptive random path selection approach for incremental learning,'' \emph{arXiv preprint arXiv:1906.01120}, 2019.

\bibitem{eeil}
F.~M. Castro, M.~J. Mar{\'\i}n-Jim{\'e}nez, N.~Guil, C.~Schmid, and K.~Alahari, ``End-to-end incremental learning,'' in \emph{Proc. Eur. Conf. Comput. Vis. (ECCV)}, 2018, pp. 233--248.

\bibitem{muc}
Y.~Liu, S.~Parisot, G.~Slabaugh, X.~Jia, A.~Leonardis, and T.~Tuytelaars, ``More classifiers, less forgetting: A generic multi-classifier paradigm for incremental learning,'' in \emph{Proc. Eur. Conf. Comput. Vis. (ECCV)}.\hskip 1em plus 0.5em minus 0.4em\relax Springer, 2020, pp. 699--716.

\bibitem{pass}
F.~Zhu, X.-Y. Zhang, C.~Wang, F.~Yin, and C.-L. Liu, ``Prototype augmentation and self-supervision for incremental learning,'' in \emph{Proc. IEEE/CVF Conf. Comput. Vis. Pattern Recognit. (CVPR)}, 2021, pp. 5871--5880.

\bibitem{wiseft}
M.~Wortsman, G.~Ilharco, J.~W. Kim, M.~Li, S.~Kornblith, R.~Roelofs, R.~G. Lopes, H.~Hajishirzi, A.~Farhadi, H.~Namkoong \emph{et~al.}, ``Robust fine-tuning of zero-shot models,'' in \emph{Proc. IEEE/CVF Conf. Comput. Vis. Pattern Recognit. (CVPR)}, 2022, pp. 7959--7971.

\bibitem{aircraft}
S.~Maji, E.~Rahtu, J.~Kannala, M.~Blaschko, and A.~Vedaldi, ``Fine-grained visual classification of aircraft,'' \emph{arXiv preprint arXiv:1306.5151}, 2013.

\bibitem{caltech}
L.~Fei-Fei, R.~Fergus, and P.~Perona, ``Learning generative visual models from few training examples: An incremental bayesian approach tested on 101 object categories,'' in \emph{Proc. IEEE/CVF Conf. Comput. Vis. Pattern Recognit. Workshops}.\hskip 1em plus 0.5em minus 0.4em\relax IEEE, 2004, pp. 178--178.

\bibitem{dtd}
M.~Cimpoi, S.~Maji, I.~Kokkinos, S.~Mohamed, and A.~Vedaldi, ``Describing textures in the wild,'' in \emph{Proc. IEEE/CVF Conf. Comput. Vis. Pattern Recognit. (CVPR)}, 2014, pp. 3606--3613.

\bibitem{eurosat}
P.~Helber, B.~Bischke, A.~Dengel, and D.~Borth, ``Eurosat: A novel dataset and deep learning benchmark for land use and land cover classification,'' \emph{IEEE J. Sel. Top. Appl. Earth Observ. Remote Sens.}, vol.~12, no.~7, pp. 2217--2226, 2019.

\bibitem{flower}
M.-E. Nilsback and A.~Zisserman, ``Automated flower classification over a large number of classes,'' in \emph{Proc. 6th Indian Conf. Comput. Vis. Graph. Image Process.}\hskip 1em plus 0.5em minus 0.4em\relax IEEE, 2008, pp. 722--729.

\bibitem{food}
L.~Bossard, M.~Guillaumin, and L.~Van~Gool, ``Food-101--mining discriminative components with random forests,'' in \emph{Proc. Eur. Conf. Comput. Vis. (ECCV)}.\hskip 1em plus 0.5em minus 0.4em\relax Springer, 2014, pp. 446--461.

\bibitem{mnist}
L.~Deng, ``The mnist database of handwritten digit images for machine learning research [best of the web],'' \emph{IEEE Signal Process. Mag.}, vol.~29, no.~6, pp. 141--142, 2012.

\bibitem{oxfordpet}
O.~M. Parkhi, A.~Vedaldi, A.~Zisserman, and C.~Jawahar, ``Cats and dogs,'' in \emph{Proc. IEEE/CVF Conf. Comput. Vis. Pattern Recognit. (CVPR)}.\hskip 1em plus 0.5em minus 0.4em\relax IEEE, 2012, pp. 3498--3505.

\bibitem{cars}
J.~Krause, M.~Stark, J.~Deng, and L.~Fei-Fei, ``3d object representations for fine-grained categorization,'' in \emph{Proc. IEEE Int. Conf. Comput. Vis. (ICCV) Workshops}, 2013, pp. 554--561.

\bibitem{sun}
J.~Xiao, J.~Hays, K.~A. Ehinger, A.~Oliva, and A.~Torralba, ``Sun database: Large-scale scene recognition from abbey to zoo,'' in \emph{Proc. IEEE Comput. Soc. Conf. Comput. Vis. Pattern Recognit.}\hskip 1em plus 0.5em minus 0.4em\relax IEEE, 2010, pp. 3485--3492.

\bibitem{adamw}
I.~Loshchilov and F.~Hutter, ``Decoupled weight decay regularization,'' \emph{arXiv preprint arXiv:1711.05101}, 2017.

\end{thebibliography}

%




\end{document}